\documentclass[10pt,twocolumn,letterpaper]{article}

\usepackage[pagenumbers]{cvpr} 

\usepackage{graphicx}
\usepackage{amsmath}
\usepackage{amssymb}
\usepackage{booktabs}
\usepackage[accsupp]{axessibility}  

\usepackage[pagebackref,breaklinks,colorlinks]{hyperref}

\usepackage[capitalize]{cleveref}
\crefname{section}{Sec.}{Secs.}
\Crefname{section}{Section}{Sections}
\Crefname{table}{Table}{Tables}
\crefname{table}{Tab.}{Tabs.}

\usepackage{algorithm}
\usepackage{algorithmic}
\usepackage{multirow}
\usepackage{colortbl}  
\usepackage[svgnames]{xcolor}
\usepackage{array}  
\usepackage{float}
\usepackage[marginal]{footmisc}

\usepackage{amsmath, bm}

\def\cvprPaperID{12849} 
\def\confName{CVPR}
\def\confYear{2023}

\begin{document}

\title{Towards Effective Visual Representations for Partial-Label Learning}

\author{
Shiyu Xia$^{1,*}$~~~~Jiaqi Lv$^{2,*}$~~~~Ning Xu$^{1}$~~~~Gang Niu$^{2,1}$~~~~Xin Geng$^{1,\dagger}$\\
{$^1$Southeast University~~$^2$RIKEN Center for Advanced Intelligence Project}\\
{\tt\small \{shiyu\_xia, xning, xgeng\}@seu.edu.cn, \{is.jiaqi.lv, gang.niu.ml\}@gmail.com
}
}
\maketitle

\footnote{*~Equal contributions. $\dagger$~Corresponding author.}

\begin{abstract}
Under partial-label learning (PLL) where, for each training instance, only a set of ambiguous candidate labels containing the unknown true label is accessible, contrastive learning has recently boosted the performance of PLL on vision tasks, attributed to representations learned by contrasting the same/different classes of entities.
Without access to true labels, positive points are predicted using pseudo-labels that are inherently noisy, and negative points often require large batches or momentum encoders, resulting in unreliable similarity information and a high computational overhead. 
In this paper, we rethink a state-of-the-art contrastive PLL method PiCO~\cite{wang2022pico}, inspiring the design of a simple framework termed PaPi (\textbf{Pa}rtial-label learning with a guided \textbf{P}rototyp\textbf{i}cal classifier), which demonstrates significant scope for improvement in representation learning, thus contributing to label disambiguation.
PaPi guides the optimization of a prototypical classifier by a linear classifier with which they share the same feature encoder, thus explicitly encouraging the representation to reflect visual similarity between categories.
It is also technically appealing, as PaPi requires only a few components in PiCO with the opposite direction of guidance, and directly eliminates the contrastive learning module that would introduce noise and consume computational resources.
We empirically demonstrate that PaPi significantly outperforms other PLL methods on various image classification tasks.

\end{abstract}

\section{Introduction}
\label{sec:intro}

\begin{figure}[ht]
  \centering
  \includegraphics[scale=0.30]{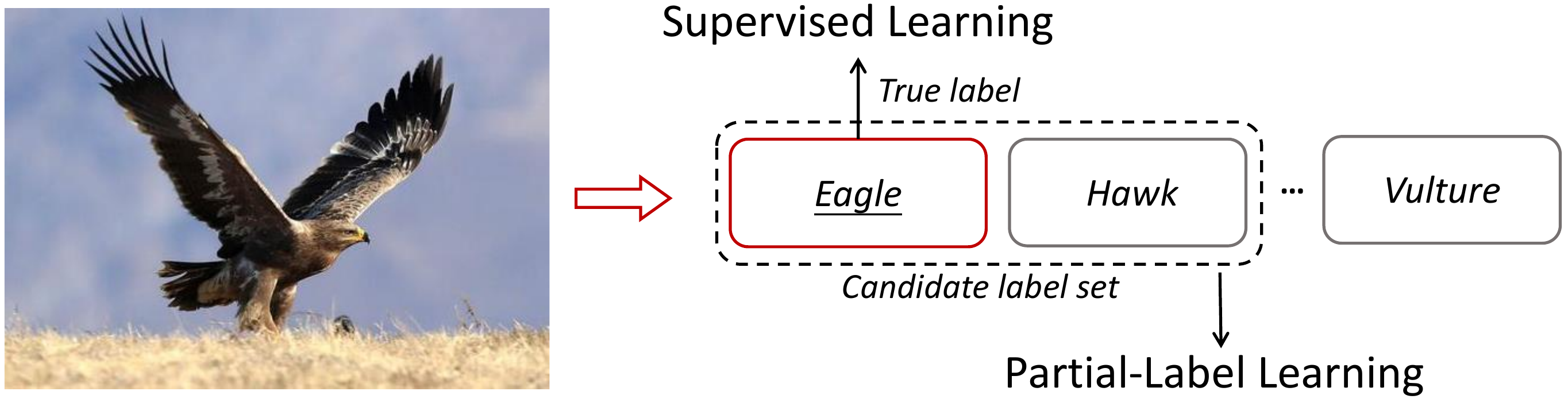}
  \caption{An image of ``Eagle'' in PLL is equipped with a candidate label set.
  PLL learns from such ambiguous supervision, in contrast to its supervised counterpart where only the true label is chosen.}
  \label{fig_1}
\end{figure}

The excellent performance of modern deep neural networks (DNNs) is attributed to the large-scale fully supervised training data, but the requirement for high-quality data poses a challenge for the practical application.
As a result, non-expert but cheap labelers are often resorted as an appealing substitute, which inevitably leads to low-quality labeled data due to their expertise limitation.
A typical situation is that the labelers have difficulty in making an accurate judgement about an instance from multiple ambiguous labels, and therefore choose multiple likely ones.
For example, in Fig.~\ref{fig_1}, it can be difficult for labelers without specialist knowledge to identify an Eagle from a Hawk, so both ``Eagle'' and ``Hawk'' are labeled as possible candidates for the true label.
Learning from such training instances with a set of possible candidate labels, where only one fixed but unknown is true, is known as \emph{partial-label learning} (PLL)~\cite{cour2011learning,zhang2015solving,yu2017maximum,xu2019partial,lv2020progressive,feng2020provably,wangopll2020,wanglfmdlp2020,wang2022pico}.
This problem naturally arises in various important applications in the real world such as web mining~\cite{luo2010learning} and image annotation~\cite{chen2017learning}.

\begin{figure*}[ht]
  \centering
  \begin{subfigure}{0.245\linewidth}
    \includegraphics[scale=0.185]{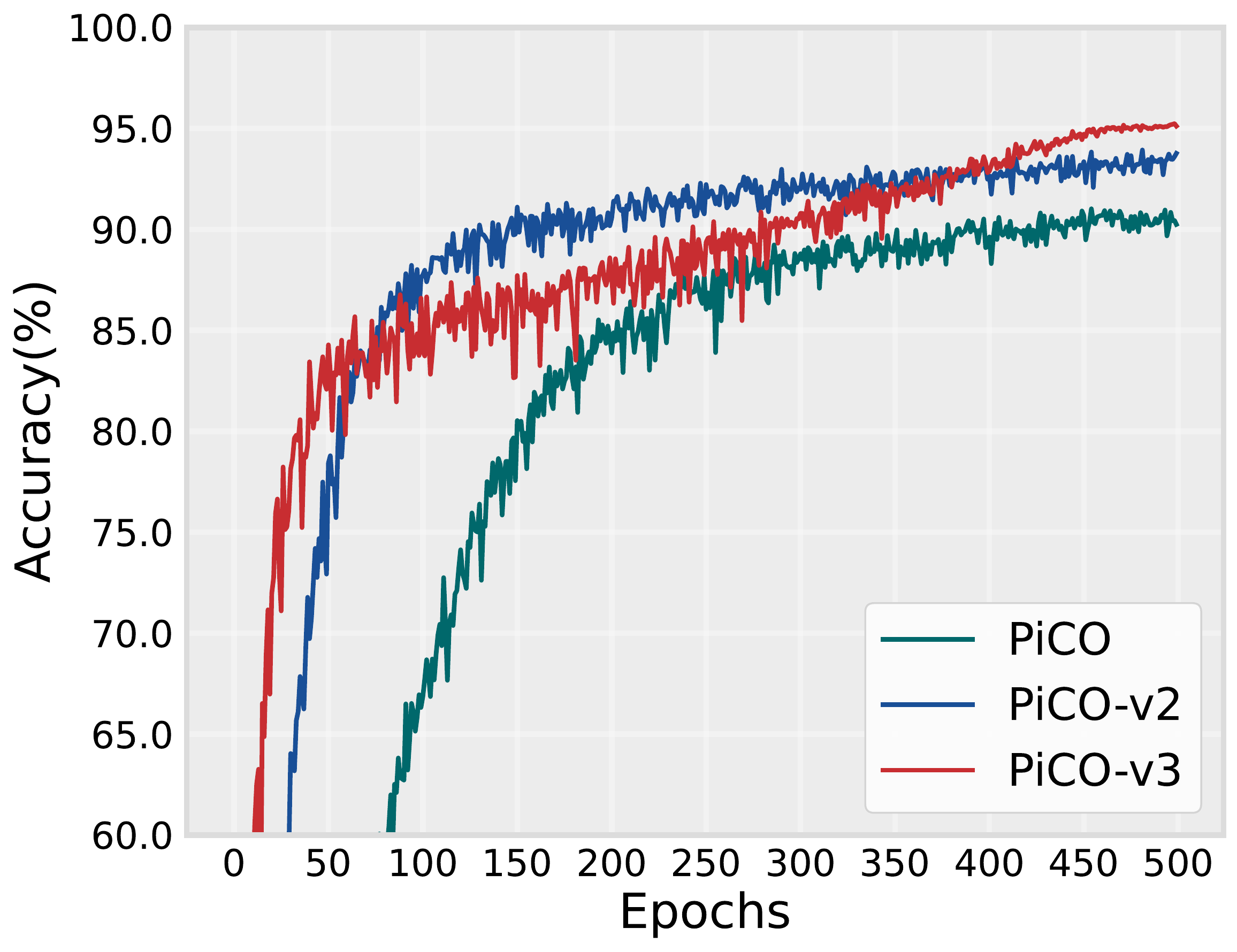}
    \caption{CIFAR-10 ($q$ = 0.7)}
    \label{fig:intro-1}
  \end{subfigure}
  \hfill
  \begin{subfigure}{0.245\linewidth}
    \includegraphics[scale=0.185]{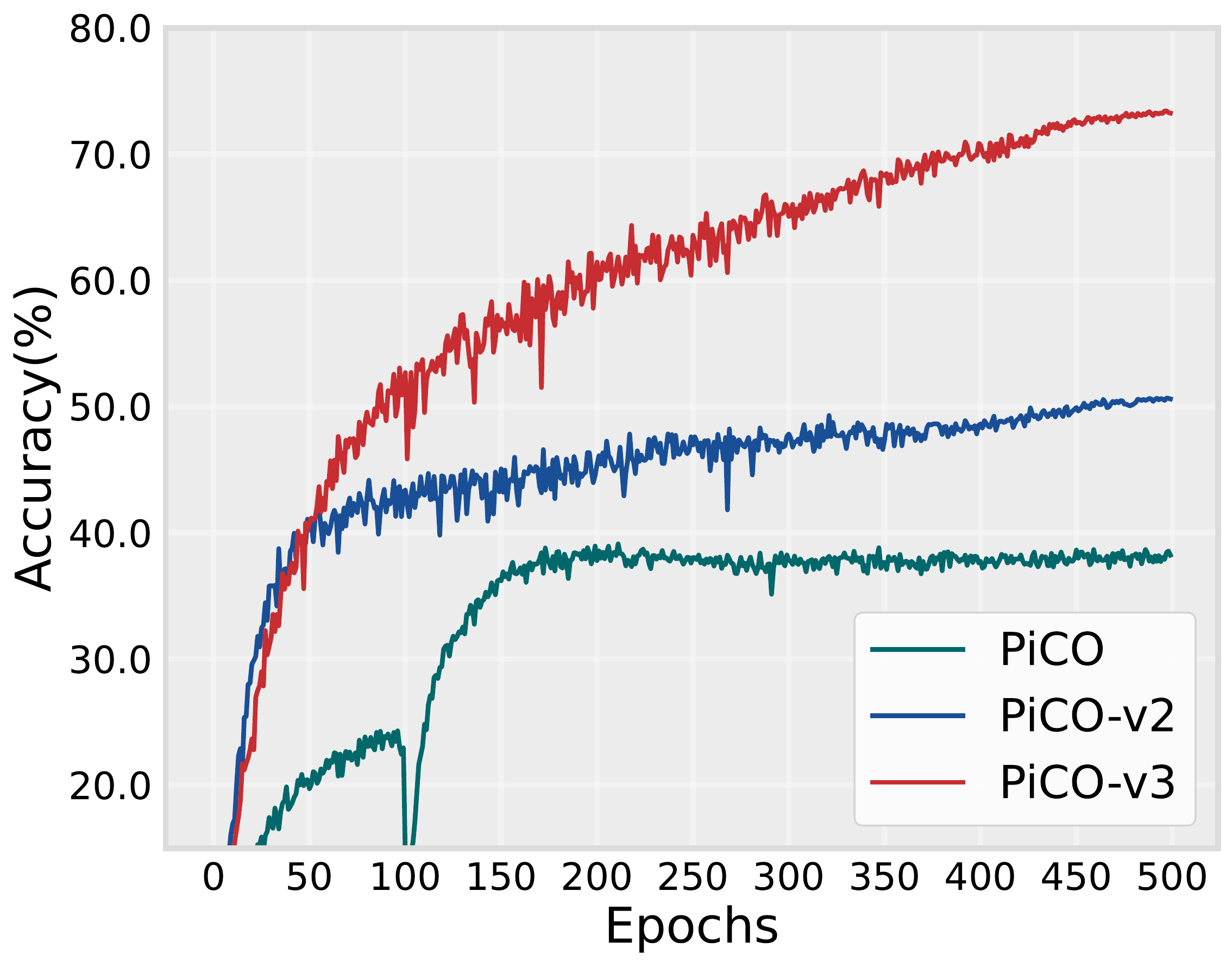}
    \caption{CIFAR-100 ($q$ = 0.2)}
    \label{fig:intro-2}
  \end{subfigure}
  \hfill
  \begin{subfigure}{0.245\linewidth}
    \includegraphics[scale=0.185]{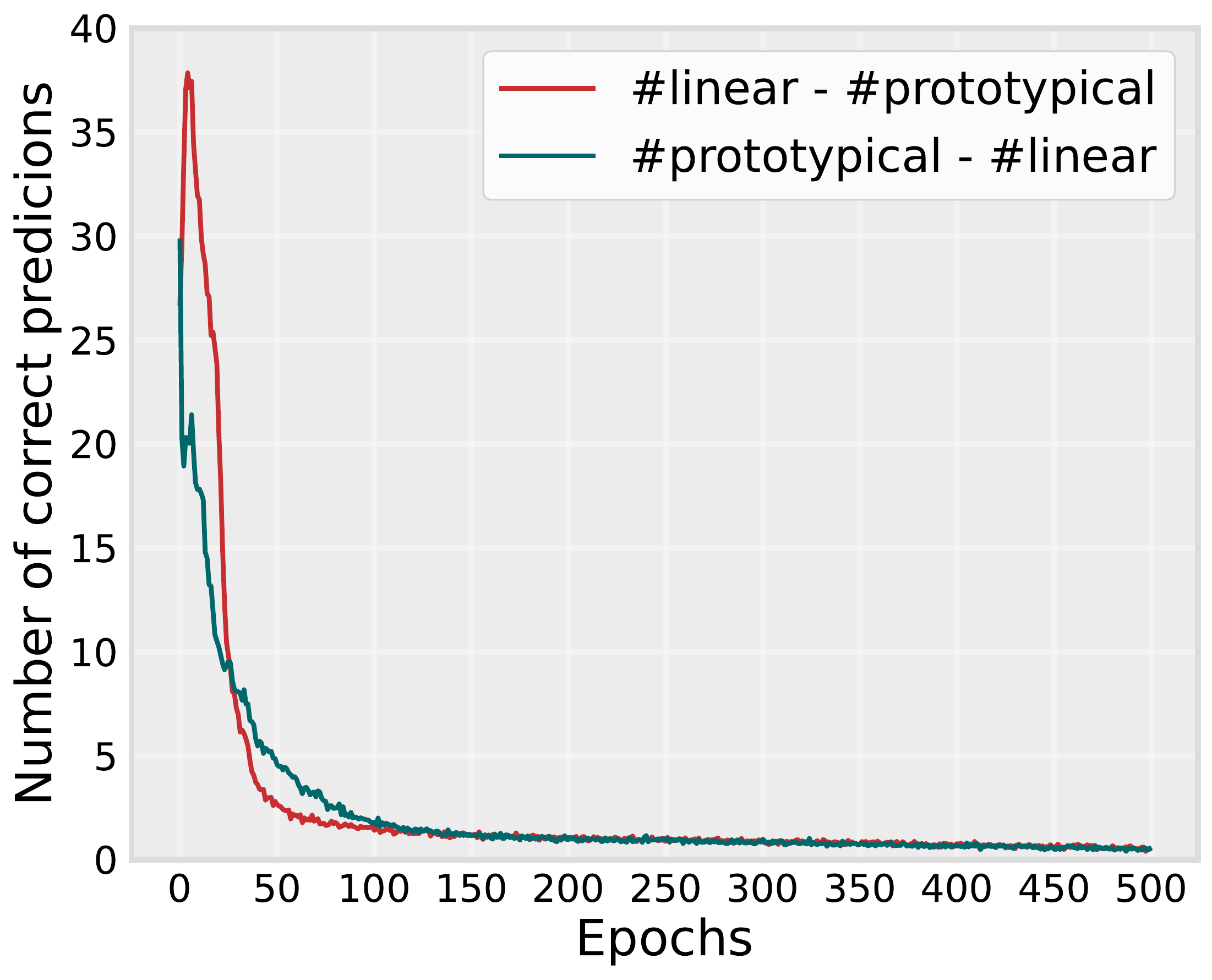}
    \caption{CIFAR-10 ($q$ = 0.7)}
    \label{fig:intro-3}
  \end{subfigure}
  \hfill
  \begin{subfigure}{0.245\linewidth}
    \includegraphics[scale=0.185]{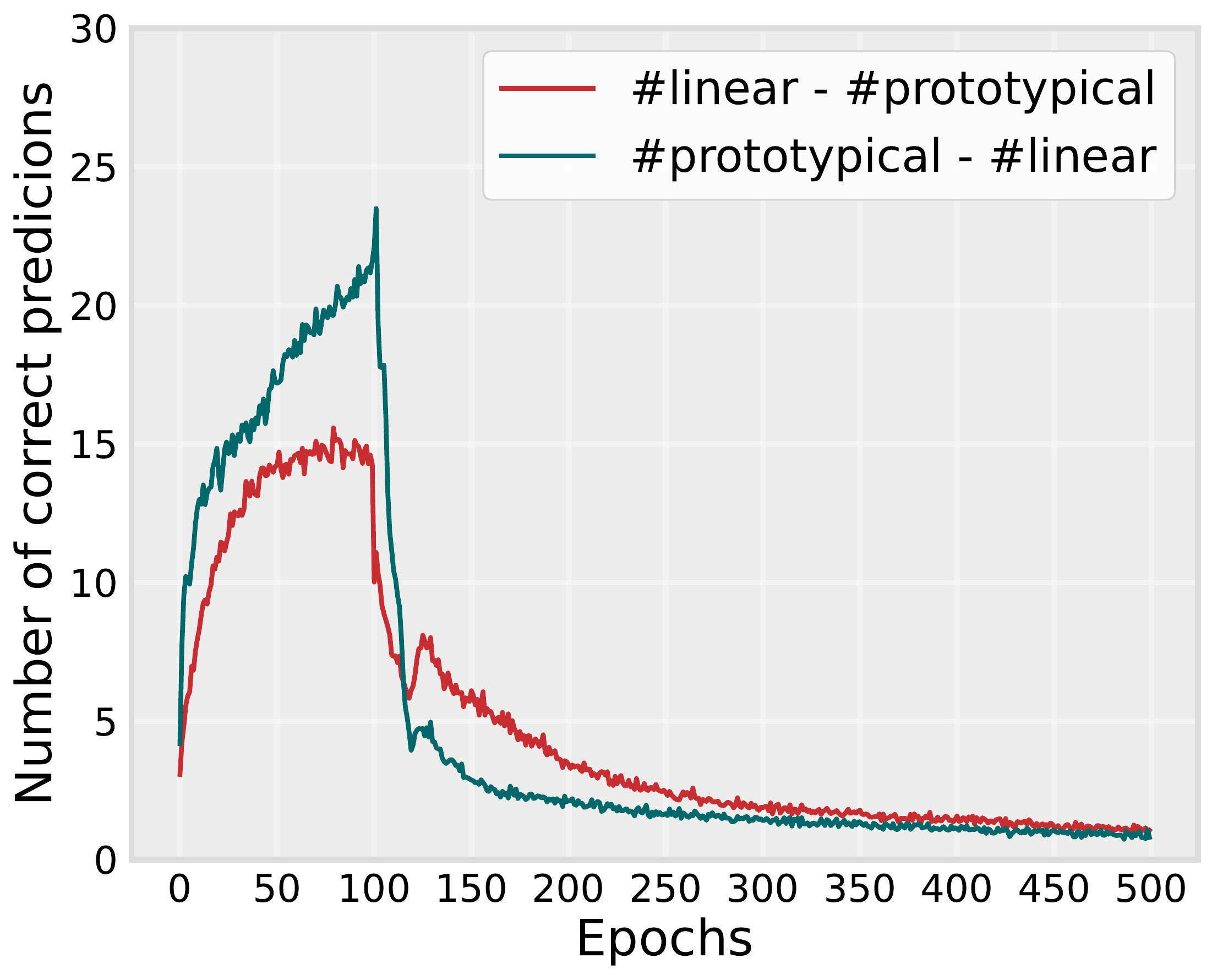}
    \caption{CIFAR-100 ($q$ = 0.2)}
    \label{fig:intro-4}
  \end{subfigure}
  \caption{Visualizations of impacts of the unreliability of pseudo positives and the improper direction of disambiguation guidance in PiCO.
  In (a)-(b), PiCO-v2 means positives are selected based on fully supervised information, \ie, true labels are known by the contrastive learning module.
  Further, PiCO-v3 removes the guidance of prototypical classifier to linear classifier, such that the linear classifier performs self-teaching.
  The red lines in (c)-(d) indicate the number of samples that were correctly classified by linear classifier and incorrectly classified by prototypical classifier per mini-batch, and the green lines are the opposite.
  The first 100 epochs shown in (d) are in a warm-up period.
  $q$ means the flipping probability of each incorrect label, which will be introduced in Sec.~\ref{sec:overview}.
  }
  \label{fig:intro_result1}
\end{figure*}

Research into PLL dates back some twenty years and a number of practical approaches have been proposed, which can be divided into identification-based strategies~\cite{jin2002learning, Zhangml2016fad, tangz2017DPLL, yu2017maximum, xu2019partial} and average-based strategies~\cite{Eyke2006, cour2011learning}, depending on how they treat candidate labels.
Recently, DNNs bring the research of PLL into a new era~\cite{yao2020deep, yao2020network, feng2020provably, lv2020progressive, wen2021leveraged, wu2021LPPL, wu2022DPLL}, among which PiCO~\cite{wang2022pico} has achieved state-of-the-art performance on multiple benchmarks.
It introduces a \emph{contrastive learning module} into PLL that uses predictions of one \emph{linear classifier} to select pseudo positive samples for each anchor point and maintains a queue of negative samples.
Meanwhile, a momentum encoder is used to improve consistency.
In addition, PiCO adds a \emph{prototypical classifier module} (called prototype-based in the original) to guide the update of the linear classifier, which is based on the idea that there exists an embedding space where points from the same class cluster around its prototype~\cite{snell2017prototypical}.
PiCO claims credit for its success to the mutual benefit of contrastive learning and prototype-based label disambiguation.

In this paper, we rethink the two modules in PiCO and empirically point out that they do not work as well in practice as one might think due to the unreliability of the pseudo positives and the improper direction of disambiguation guidance.
Fig.~\ref{fig:intro-1} and Fig.~\ref{fig:intro-2} show accuracy of three versions of PiCO.
Fig.~\ref{fig:intro-3} and Fig.~\ref{fig:intro-4} show the performance differences between the linear and prototypical classifier during training.
Fig.~\ref{fig:intro_result1} delivers two important messages: 
(1) noisy pseudo-labels do lead to significant performance degradation, and (2) the phenomenon ``poor teacher teaches good student'' possibly happens.
Specifically, the good student, the linear classifier, always made more correct predictions than its teacher, the prototypical classifier, at the beginning. 
In some cases, due to the forced direction of guidance, the teacher performed better than the student for a while, but soon the teacher had nothing new to teach the student, shown in Fig.~\ref{fig:intro-3}.
And sometimes the student's advantage was even maintained until convergence as shown in Fig.~\ref{fig:intro-4}.
These also explain the significant improvement compared to PiCO-v2 after PiCO-v3 made the clever student perform self-teaching.

Inspired by the above observations, we propose a simple PLL framework termed \textbf{PaPi}, \ie, \emph{\textbf{Pa}rtial-label learning with a guided \textbf{P}rototyp\textbf{i}cal classifier}.
PaPi directly eliminates the contrastive learning module which introduces noisy positives, and adopts the opposite direction of disambiguation guidance compared to PiCO.
Specifically, PaPi produces a similarity distribution over classes for each sample based on a softmax function over distances to class-specific prototypes in a projected low-dimensional space.
Afterwards, PaPi aligns the distribution with the disambiguated probability post-processed from one linear classifier prediction.
Meanwhile, the linear classifier performs self-teaching wherein each stage of learning is guided by the current and previous stages.
We conduct extensive experiments on multiple image classification tasks.
PaPi surpasses the state-of-the-art PLL methods by a large margin, with a $\bm{4.57\%}$ improvement on CIFAR-100 in very difficult scenarios.
Moreover, PaPi learns effective representations efficiently without using neither large batches nor a momentum encoder, where training instances from the same class are grouped into tighter clusters.
Our main \textbf{contributions} are summarized as follows:
\begin{itemize}
\item We propose a simple PLL framework termed PaPi which explicitly encourages the representation to reflect visual similarity between categories, such that PaPi is remarkable for improving the class-level discrimination of learned representation.
\item Extensive experiments on various image classification datasets with different generation processes of candidate labels demonstrate PaPi significantly outperforms state-of-the-art PLL methods.
\end{itemize}

\section{Related Work}
\label{sec:relat}

\begin{figure*}[ht]
\centering
\includegraphics[scale=0.51]{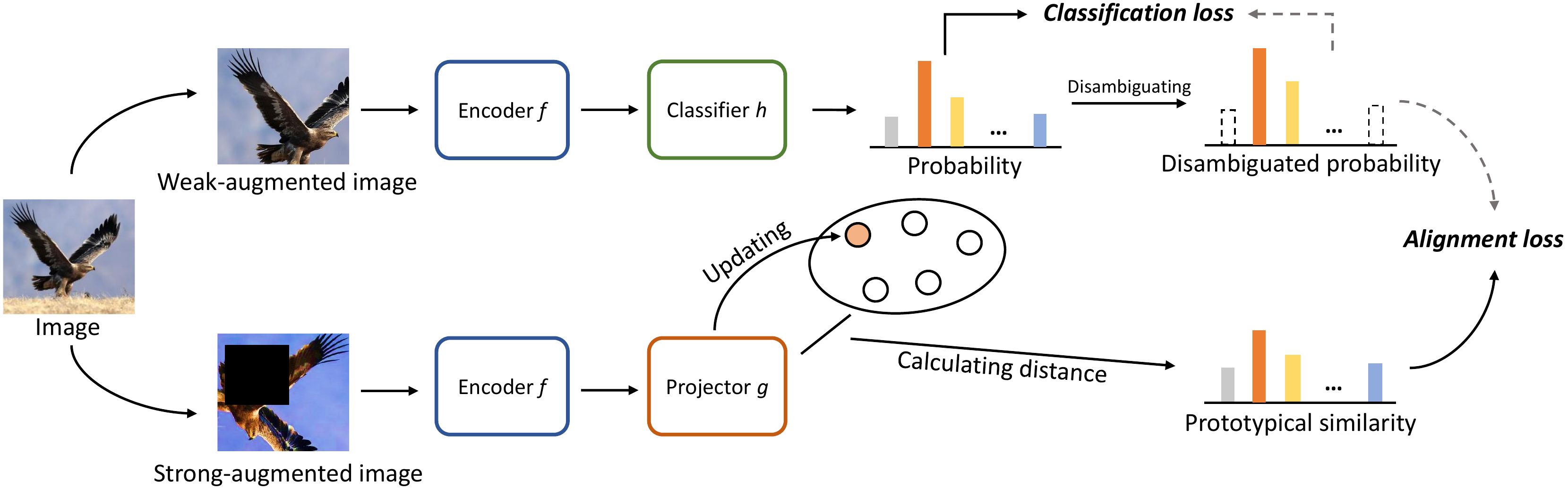}
\caption{Illustration of PaPi. 
With a prototypical alignment loss term, we reduce the diversity between a prototypical similarity distribution over classes for each sample based on a softmax function over distances to class-specific prototypes in a projected low-dimensional space and the disambiguated probability post-processed from one linear classifier. 
Using a cross-entropy loss, we learn the linear classifier in a self-teaching fashion. Note that the dotted gray arrow indicates the stop gradient operation.}
\label{fig_pipeline}
\end{figure*}

\subsection{Partial-Label Learning}
A plethora of PLL works purify the candidate labels with a common goal to select the most likely true label in the training phase, which are named as \emph{identification-based} methods~\cite{Zhangml2016fad, tangz2017DPLL, lyu2021gmpll}.
For this purpose, a maximum likelihood model is proposed in~\cite{jin2002learning} using the expectation-maximization algorithm, which pioneers this main research route.
In addition, some works~\cite{yu2017maximum, xu2019partial} treat the true label as a latent variable and identify it by leveraging the topological information from feature space.
In contrast to this research route, \emph{average-based} methods~\cite{Eyke2006, cour2011learning, zhang2015solving} treat all the candidate labels equally and the prediction is made by averaging their modeling outputs.
However, most of the above methods need elaborately designed learning objectives and specific solutions, which make them inefficient for scaling up to large scale datasets.
With the advances in the deep learning era, PLL has been broadly explored in recent works~\cite{yao2020deep, yao2020network, feng2020provably, lv2020progressive, wu2021LPPL, wang2022pico, wu2022DPLL, xia2022ambiguity}.
PRODEN~\cite{lv2020progressive} accomplishes the update of the model and identification of true labels seamlessly along with a classifier-consistent risk estimator.
Two methods which are risk-consistent and classifier-consistent are derived based on the uniform partial labels generation process~\cite{feng2020provably}.
VALEN~\cite{xu2021instance} applies label enhancement~\cite{xu2020variational, xu2019label} to iteratively recover label distribution for each instance, where instance-dependent PLL is firstly explored.
\cite{lv2021robustness} proposes a general scenario called unreliable PLL where the ground-truth label can be unnecessarily included in the candidate label set.
Consistency regularization is revisited in deep PLL~\cite{wu2022DPLL}, which employs regularization on candidate labels based on specific data augmentation techniques.
In this paper, we rethink a state-of-the-art method PiCO~\cite{wang2022pico} and propose a simple framework named PaPi, which demonstrates significant scope for improvement in representation learning, thus contributing to label disambiguation.
Moreover, current methods cannot effectively handle both uniform and instance-dependent PLL setting, especially when they are confronted with \emph{high ambiguity levels} and \emph{instance-dependent ambiguity}.

\subsection{Prototype Learning}
Prototype learning (PL) learns a representation space where instances are imposed to be closer to its corresponding class prototype.
PL has demonstrated robust performance in few-shot learning~\cite{snell2017prototypical, pahde2021multimodal}, zero-shot learning~\cite{xu2020attribute}, noisy-label learning~\cite{li2020mopro}, semi-supervised learning~\cite{han2020unsupervised}, class-imbalanced learning~\cite{wei2022prototypical}, contrastive learning~\cite{li2020prototypical}, \etc.
For example, MoPro~\cite{li2020mopro} aims to learn an embedding space where instances from the same class gather around its class prototype, which is achieved with a prototypical contrastive loss and an instance contrastive loss.

\section{The Proposed Framework}
\label{sec:method}
In this section, we firstly introduce the preliminaries in Sec.~\ref{sec:overview}. 
Then we present our PaPi in detail in Sec.~\ref{sec:align}.
Fig.~\ref{fig_pipeline} gives a brief illustration of PaPi.

\subsection{Preliminaries}
\label{sec:overview}

\paragraph{Notations.}
Let $\mathcal{X}$ be the input space, $\mathcal{Y}$ = $\{1,2,...,K\}$ be the label space with $K$ class labels.
One of the key differences between PLL and supervised learning is that the latent ground-truth label $y_i\in\mathcal{Y}$ of an instance $\boldsymbol{x}_i$ is not provided but always included in a candidate label set $Y_i\subseteq\mathcal{Y}$.
Following~\cite{wang2022pico}, we generate uniform partial labels by flipping each incorrect label $\Bar{y}_i\neq{y}_i$ with a probability $q=P(\Bar{y}_i\in Y_i|\Bar{y}_i\neq{y}_i)$.
For generating instance-dependent partial labels, we follow the same procedure in~\cite{xu2021instance}. 
We set the flipping probability of each incorrect label $j$ as $q_{j}(\boldsymbol{x}_i)=\frac{\hat{g}_j(\boldsymbol{x}_i)}{\max_{k\in \Bar{Y}_i}\hat{g}_k(\boldsymbol{x}_i)}$ with a pre-trained DNN $\hat{g}$, where $\Bar{Y}_i$ is the incorrect label set of $\boldsymbol{x}_i$.
Given a PLL training dataset $\mathcal{D} = \{(\boldsymbol{x}_i, Y_i) \vert 1 \leq i \leq N \}$, the goal of PLL is to learn a mapping function which can predict the true label related to the unseen inputs.

\paragraph{Overview.}
For each sample $(\boldsymbol{x}_i, Y_i)$, we generate one weakly-augmented view $(aug_{1}({\boldsymbol{x}}_{i}), Y_i)$ and one strongly-augmented view $(aug_{2}({\boldsymbol{x}}_{i}), Y_i)$, where $aug_{1}(\cdot)$ and $aug_{2}(\cdot)$ represent weak and strong augmentation function respectively.
Then both augmented samples are separately fed into the weight-shared encoder network $f(\cdot)$, yielding a pair of representations $\boldsymbol{v}^{1}_{i} = f(aug_{1}({\boldsymbol{x}}_{i}))$ and $\boldsymbol{v}^{2}_{i} = f(aug_{2}({\boldsymbol{x}}_{i}))$.
In order to improve the representation quality~\cite{chen2020simple, grill2020bootstrap}, we further map $\boldsymbol{v}^{1}_{i}$ to $\boldsymbol{z}^{1}_{i} = g(\boldsymbol{v}^{1}_{i}) \in \mathbb{R}^{d_p}$ by utilizing a projector network $g(\cdot)$, which is also operated on $\boldsymbol{v}^{2}_{i}$ to obtain $\boldsymbol{z}^{2}_{i}$.
Therefore, for each instance $\boldsymbol{x}_{i}$, the corresponding low-dimensional representation $\boldsymbol{z}_{i}$ consists of $\boldsymbol{z}^{1}_{i}$ and $\boldsymbol{z}^{2}_{i}$.
Note that $\boldsymbol{z}^{1}_{i}$ and $\boldsymbol{z}^{2}_{i}$ are further normalized to the unit sphere in $\mathbb{R}^{d_p}$, based on which we update class-specific prototypes $C \in \mathbb{R}^{K \times d_p}$ in a moving-average fashion.
We call $f(\cdot)$ and $g(\cdot)$ as representor when the context is clear.
Meanwhile, the classifier $h(\cdot)$ receives $\boldsymbol{v}^{1}_{i}$ as input and outputs $\boldsymbol{r}_{i}=h(\boldsymbol{v}^{1}_{i})$.
Besides the classification loss term implemented by a cross-entropy loss, we propose a prototypical alignment loss term which reduces the diversity between a similarity distribution over classes based on a softmax function over distances to class-specific prototypes and the probability disambiguated from the linear classifier prediction.
Based on the feature encoder shared by linear classifier and projector, we build the connection between the effective representation space and the label disambiguation to facilitate the model training.

\subsection{PLL with A Guided Prototypical Classifier}
\label{sec:align}
As we mentioned earlier, PaPi directly eliminates the contrastive learning module and adopts the opposite direction of disambiguation guidance compared to PiCO.
Specifically, PaPi introduces a prototypical alignment loss term which guides the optimization of a prototypical classifier.

\paragraph{Prototypical alignment.}
To begin with, we describe prototypical alignment loss term.
Given each sample $(\boldsymbol{x}_i, Y_i)$, we produce a prototypical similarity distribution $\boldsymbol{s}^{l}_{i}$ over classes by a softmax function over distances to class-specific prototypes:
\begin{align}\label{Eq.1}
    {s}^{l}_{ij} = \frac{\exp(\boldsymbol{z}^{l}_{i}\cdot\boldsymbol{c}_{j}/\tau)}{\sum_{k=1}^{K}\exp(\boldsymbol{z}^{l}_{i}\cdot \boldsymbol{c}_{k}/\tau)},
\end{align}
where $l \in \{1,2\}$ denotes different augmentation views, $\tau$ is the temperature and we use cosine similarity as distance measure.
Meanwhile, we update the disambiguated probability based on the linear classifier prediction via putting more mass on more possible candidate labels:
\begin{align}
    u_{ij}=\left\{
    \begin{aligned}
    &\quad\frac{r_{ij}}{\sum\nolimits_{l\in {Y}_i} r_{il}}  & \text{if }~~j \in {Y}_i,\\
    &\quad\quad\quad 0\quad\quad & \text{otherwise}.
    \end{aligned}
    \right.
\end{align}
where $j$ denotes the indices of candidate labels. 
Due to the low-quality prediction in initial training stage, we further adopt a soft and moving-average mechanism to update the probability, which also stabilizes the training process:
\begin{align}\label{Eq.3}
    \boldsymbol{p}_{i} = \lambda\boldsymbol{p}_{i} + (1 - \lambda)\boldsymbol{u}_{i}, 
\end{align}
where $\lambda$ is a balancing factor.
Intuitively speaking, initializing the disambiguated probability $\boldsymbol{p}_{i}$ as a uniform distribution leads a good start as the representations are less distinguishable at the beginning.
Therefore, we set $p_{ij} = 1/|{Y}_i|$ if $j \in {Y}_i$, otherwise $p_{ij} = 0$.
Now we implement the per-sample alignment loss term by minimizing the Kullback-Leibler (KL) divergence of the prototypical similarity distribution $\boldsymbol{s}^{l}_{i}$ and the disambiguated probability $\boldsymbol{p}_{i}$:
\begin{align}
    \mathcal{L}^{i}_{pa}(\boldsymbol{p}_{i}, \boldsymbol{s}^{l}_{i}) = \sum\nolimits_{l\in \{1,2\}}D_{KL}(\boldsymbol{p}_{i}||\boldsymbol{s}^{l}_{i}).
\end{align}

\begin{algorithm}[tb]
\renewcommand{\algorithmicrequire}{ \textbf{Input:}} 
\renewcommand{\algorithmicensure}{ \textbf{Output:}} 
\caption{Pseudo-code of PaPi.}
\label{alg:algorithm1}
\begin{algorithmic}[1] 
\REQUIRE Training dataset $\mathcal{D}$, encoder $f(\cdot)$, projector $g(\cdot)$, classifier $h(\cdot)$, mini-batch size $B$, epochs $T_{max}$, hyper-parameters $\lambda, \phi, \alpha, \gamma$.
\FOR{$t=1,2,...,T_{max}$}
\STATE Shuffle $\mathcal{D}$ into $\frac{|\mathcal{D}|}{B}$ mini-batches;
\FOR{$b=1,2,...,\frac{|\mathcal{D}|}{B}$}
\STATE Obtain the disambiguation target by Eq.\eqref{Eq.3};
\STATE Create new training samples by Eq.\eqref{Eq.5};
\STATE Generate prototypical similarity distribution for these new samples by Eq.\eqref{Eq.1};
\STATE Update parameter of $f(\cdot)$, $g(\cdot)$ and $h(\cdot)$ by minimizing the empirical risk with loss in Eq.\eqref{Eq.9};
\STATE Update prototypes by Eq.\eqref{Eq.7};
\ENDFOR
\ENDFOR
\ENSURE parameter of $f(\cdot)$ and $h(\cdot)$.
\end{algorithmic}
\end{algorithm}

Considering that the label ambiguity generally exists in the training process, we further regularize the model from memorizing such ambiguous labels to learn more robust representations inspired by Mixup~\cite{zhang2017mixup}.
Specifically, we construct new training samples by linearly interpolating a sample $($indexed by $i$$)$ with another sample $($indexed by $m(i)$$)$ obtained from the same mini-batch randomly:
\begin{align}\label{Eq.5}
    \hat{\boldsymbol{x}}_{i} = \phi\boldsymbol{x}_{i} + (1 - \phi)\boldsymbol{x}_{m(i)}, 
\end{align}
where $\phi \sim Beta(\alpha, \alpha)$ and $\alpha$ is a hyperparameter.
Denoting $\hat{\boldsymbol{s}}^{l}_{i}$ as the prototypical similarity distribution for $\hat{\boldsymbol{x}}_{i}$, we define the per-sample augmented alignment loss term as a weighted combination of two original loss term with respect to $\boldsymbol{p}_{i}$ and $\boldsymbol{p}_{m(i)}$:
\begin{align}
    \mathcal{L}^{i}_{ali} = \phi \mathcal{L}^{i}_{pa}(\boldsymbol{p}_{i}, \hat{\boldsymbol{s}}^{l}_{i}) + (1-\phi) \mathcal{L}^{i}_{pa}(\boldsymbol{p}_{m(i)}, \hat{\boldsymbol{s}}^{l}_{i}).
\end{align}

\begin{table*}[ht]
	\centering
	\setlength{\tabcolsep}{3.8mm}{
		\begin{tabular}{c|ccccc}
			\toprule[1.5pt]
			Dataset & Method & $q$ = 0.1 & $q$ = 0.3 & $q$ = 0.5 & $q$ = 0.7\\
			\midrule
			\multirow{8}*{Fashion-MNIST} & \cellcolor{Gainsboro} Supervised &  \multicolumn{4}{c}{\cellcolor{Gainsboro} 96.41 $\pm$ 0.02\%} \\
		    ~ & PaPi (ours) & \textbf{95.79 $\pm$ 0.16\%} & \textbf{95.63 $\pm$ 0.16\%} & \textbf{95.60 $\pm$ 0.07\%} & \textbf{95.16 $\pm$ 0.10\%} \\
            ~ & DPLL & 95.76 $\pm$ 0.11\% & 95.61 $\pm$ 0.15\% & 95.52 $\pm$ 0.08\% & 94.95 $\pm$ 0.07\% \\
            ~ & PiCO & 94.89 $\pm$ 0.02\% & 94.35 $\pm$ 0.09\% & 93.87 $\pm$ 0.16\% & 90.05 $\pm$ 2.04\% \\
            ~ & PRODEN & 94.62 $\pm$ 0.25\% & 94.40 $\pm$ 0.08\% & 93.92 $\pm$ 0.27\% & 93.62 $\pm$ 0.28\% \\
            ~ & LWS & 94.73 $\pm$ 0.15\% & 94.50 $\pm$ 0.10\% & 93.16 $\pm$ 0.74\% & 91.08 $\pm$ 0.48\% \\
            ~ & RC & 95.27 $\pm$ 0.26\% & 95.02 $\pm$ 0.07\% & 94.73 $\pm$ 0.14\% & 93.92 $\pm$ 0.37\% \\
            ~ & CC & 95.30 $\pm$ 0.22\% & 94.95 $\pm$ 0.16\% & 94.64 $\pm$ 0.14\% & 93.42 $\pm$ 0.24\% \\
			\midrule
			\multirow{8}*{SVHN} & \cellcolor{Gainsboro} Supervised &  \multicolumn{4}{c}{\cellcolor{Gainsboro} 97.64 $\pm$ 0.03\%} \\
		    ~ & PaPi (ours) & \textbf{97.57 $\pm$ 0.04\%} & \textbf{97.48 $\pm$ 0.07\%} & \textbf{97.52 $\pm$ 0.04\%} & \textbf{97.29 $\pm$ 0.07\%} \\
            ~ & DPLL & 97.27 $\pm$ 0.06\% & 97.16 $\pm$ 0.07\% & 96.82 $\pm$ 0.08\% & 95.14 $\pm$ 0.10\% \\
            ~ & PiCO & 95.71 $\pm$ 0.12\% & 95.48 $\pm$ 0.16\% & 94.58 $\pm$ 0.33\% & 94.15 $\pm$ 0.12\% \\
            ~ & PRODEN & 96.82 $\pm$ 0.03\% & 96.61 $\pm$ 0.10\% & 96.60 $\pm$ 0.21\% & 95.49 $\pm$ 0.28\% \\
            ~ & LWS & 96.37 $\pm$ 0.01\% & 96.32 $\pm$ 0.43\% & 34.96 $\pm$ 0.15\% & 18.37 $\pm$ 1.12\% \\
            ~ & RC & 97.11 $\pm$ 0.10\% & 97.13 $\pm$ 0.04\% & 97.08 $\pm$ 0.06\% & 96.33 $\pm$ 0.05\% \\
            ~ & CC & 97.02 $\pm$ 0.03\% & 96.96 $\pm$ 0.05\% & 96.79 $\pm$ 0.10\% & 96.24 $\pm$ 0.12\% \\
      \midrule
			\multirow{8}*{CIFAR-10} & \cellcolor{Gainsboro} Supervised &  \multicolumn{4}{c}{\cellcolor{Gainsboro} 97.36 $\pm$ 0.04\%} \\
		    ~ & PaPi (ours) & \textbf{97.33 $\pm$ 0.06\%} & \textbf{97.26 $\pm$ 0.08\%} & \textbf{96.90 $\pm$ 0.09\%} & \textbf{96.58 $\pm$ 0.07\%} \\
            ~ & DPLL & 95.85 $\pm$ 0.22\% & 95.57 $\pm$ 0.15\% & 95.31 $\pm$ 0.11\% & 94.08 $\pm$ 0.16\% \\
            ~ & PiCO & 95.09 $\pm$ 0.31\% & 94.52 $\pm$ 0.39\% & 93.97 $\pm$ 0.55\% & 92.88 $\pm$ 0.57\% \\
            ~ & PRODEN & 93.85 $\pm$ 0.60\% & 93.36 $\pm$ 0.53\% & 92.95 $\pm$ 0.37\% & 90.98 $\pm$ 0.28\% \\
            ~ & LWS & 91.33 $\pm$ 0.08\% & 90.17 $\pm$ 0.24\% & 65.54 $\pm$ 1.64\% & 44.73 $\pm$ 1.43\% \\
            ~ & RC & 94.84 $\pm$ 0.38\% & 94.64 $\pm$ 0.07\% & 93.11 $\pm$ 0.03\% & 86.11 $\pm$ 2.85\% \\
            ~ & CC & 94.65 $\pm$ 0.27\% & 94.31 $\pm$ 0.20\% & 92.48 $\pm$ 0.33\% & 89.78 $\pm$ 0.35\% \\
      \midrule
      \multirow{8}*{CIFAR-100-H} & \cellcolor{Gainsboro} Supervised &  \multicolumn{4}{c}{\cellcolor{Gainsboro}  82.31 $\pm$ 0.06\%} \\
            ~ & PaPi (ours) & \textbf{79.54 $\pm$ 0.13\%} & \textbf{79.38 $\pm$ 0.19\%} & \textbf{78.75 $\pm$ 0.17\%} & \textbf{76.25 $\pm$ 0.24\%} \\
                ~ & DPLL & 75.71 $\pm$ 0.12\% & 74.55 $\pm$ 0.09\% & 73.98 $\pm$ 0.12\% & 71.21 $\pm$ 0.18\% \\
                ~ & PiCO & 73.81 $\pm$ 0.13\% & 72.45 $\pm$ 0.44\% & 72.03 $\pm$ 0.19\% & 68.71 $\pm$ 1.20\% \\
                ~ & PRODEN & 73.24 $\pm$ 0.19\% & 72.21 $\pm$ 0.97\% & 70.11 $\pm$ 0.14\% & 67.25 $\pm$ 0.88\% \\
                ~ & LWS & 64.77 $\pm$ 1.45\% & 44.25 $\pm$ 0.68\% & 40.23 $\pm$ 1.27\% & 33.21 $\pm$ 1.22\% \\
                ~ & RC & 72.31 $\pm$ 0.16\% & 71.78 $\pm$ 0.12\% & 69.88 $\pm$ 0.24\% & 52.14 $\pm$ 1.22\% \\
                ~ & CC & 72.57 $\pm$ 0.88\% & 71.33 $\pm$ 0.90\% & 70.22 $\pm$ 0.87\% & 65.22 $\pm$ 0.89\% \\
			\bottomrule[1.5pt]
		\end{tabular}
	}
	\caption{Classification accuracy (mean $\pm$ std) on Fashion-MNIST, SVHN, CIFAR-10 and CIFAR-100-H with uniform partial labels under different ambiguity levels. The best accuracy is highlighted in bold.}
	\label{tab:benchmark_result1}
\end{table*}

\begin{table*}[ht]
	\centering
	\setlength{\tabcolsep}{4.0mm}{
		\begin{tabular}{c|ccccc}
			\toprule[1.5pt]
			Dataset & Method & $q$ = 0.01 & $q$ = 0.05 & $q$ = 0.1 & $q$ = 0.2\\
			\midrule
			\multirow{8}*{CIFAR-100} & \cellcolor{Gainsboro} Supervised &  \multicolumn{4}{c}{\cellcolor{Gainsboro}  82.31 $\pm$ 0.06\%} \\
		    ~ & PaPi (ours) & \textbf{82.23 $\pm$ 0.27\%} & \textbf{81.60 $\pm$ 0.18\%} & \textbf{81.65 $\pm$ 0.27\%} & \textbf{79.49 $\pm$ 0.22\%} \\
            ~ & DPLL & 79.74 $\pm$ 0.07\% & 78.97 $\pm$ 0.13\% & 78.51 $\pm$ 0.24\% & 75.77 $\pm$ 0.06\% \\
            ~ & PiCO & 73.78 $\pm$ 0.15\% & 72.78 $\pm$ 0.38\% & 71.55 $\pm$ 0.31\% & 48.76 $\pm$ 1.19\% \\
            ~ & PRODEN & 73.20 $\pm$ 0.18\% & 72.13 $\pm$ 1.10\% & 71.88 $\pm$ 0.32\% & 52.10 $\pm$ 0.67\% \\
            ~ & LWS & 64.55 $\pm$ 1.98\% & 50.19 $\pm$ 0.34\% & 44.93 $\pm$ 1.09\% & 37.97 $\pm$ 1.42\% \\
            ~ & RC & 75.36 $\pm$ 0.06\% & 74.44 $\pm$ 0.31\% & 73.79 $\pm$ 0.29\% & 64.74 $\pm$ 0.82\% \\
            ~ & CC & 76.16 $\pm$ 0.32\% & 75.04 $\pm$ 0.10\% & 73.56 $\pm$ 0.27\% & 71.43 $\pm$ 0.32\% \\
			\midrule
			\multirow{8}*{Mini-Imagenet} & \cellcolor{Gainsboro} Supervised &  \multicolumn{4}{c}{\cellcolor{Gainsboro} 74.45 $\pm$ 0.18\%} \\
		    ~ & PaPi (ours) & \textbf{72.43 $\pm$ 0.02\%} & \textbf{72.08 $\pm$ 0.12\%} & \textbf{71.54 $\pm$ 0.34\%} & \textbf{67.81 $\pm$ 0.36\%} \\
            ~ & DPLL & 72.20 $\pm$ 0.01\% & 71.45 $\pm$ 0.25\% & 70.89 $\pm$ 0.28\% & 61.19 $\pm$ 0.73\% \\
            ~ & PiCO & 67.24 $\pm$ 1.73\% & 66.44 $\pm$ 1.41\% & 64.00 $\pm$ 1.62\% & 43.46 $\pm$ 1.87\% \\
            ~ & PRODEN & 65.15 $\pm$ 0.80\% & 61.70 $\pm$ 0.68\% & 58.28 $\pm$ 0.88\% & 37.04 $\pm$ 0.72\% \\
            ~ & LWS & 65.85 $\pm$ 0.51\% & 59.39 $\pm$ 0.90\% & 54.12 $\pm$ 0.25\% & 21.62 $\pm$ 0.62\% \\
            ~ & RC & 64.60 $\pm$ 0.55\% & 59.25 $\pm$ 0.70\% & 51.11 $\pm$ 0.11\% & 22.70 $\pm$ 1.12\% \\
            ~ & CC & 64.74 $\pm$ 0.44\% & 58.92 $\pm$ 1.06\% & 49.23 $\pm$ 0.90\% & 20.73 $\pm$ 0.61\% \\
			\bottomrule[1.5pt]
		\end{tabular}
	}
	\caption{Classification accuracy (mean $\pm$ std) on CIFAR-100 and Mini-Imagenet with uniform partial labels under different ambiguity levels.
	The best accuracy is highlighted in bold.}
	\label{tab:benchmark_result2}
\end{table*}

\paragraph{Prototype evolving.}
Instead of calculating the prototype representation as the normalized mean representation after every training iteration, we update the prototype $\boldsymbol{c}_{k}$ of class $k$ as a moving-average of the normalized representations for samples with pseudo-label $k$:
\begin{align}\label{Eq.7}
    \boldsymbol{c}_{k} = \gamma\boldsymbol{c}_{k} + (1 - \gamma)\boldsymbol{z}_{j}, \quad \text{if }j \in I_{k},
\end{align}
where $I_{k}$ denotes the index set of samples pseudo-labeled with class $k$ and $\gamma$ is a balancing factor.
Note that the prototypes are further normalized to the unit sphere.

\paragraph{Label disambiguation.}
As mentioned earlier, we build the connection between the highly effective representation space and the label disambiguation based on the weight-shared encoder.
For multi-class classification, we adopt cross-entropy loss as the classification loss term with $\boldsymbol{p}_{i}$ being the learning target.
The per-sample classification loss term is defined as:
\begin{align}
    \mathcal{L}^{i}_{cla} = - \sum_{j=1}^K {p_{ij}\cdot\log r_{ij}}.
\end{align}

\paragraph{Model updating.}
Putting it all together, the overall training objective for each sample is to minimize a weighted sum of all loss terms:
\begin{align}\label{Eq.9}
    \mathcal{L}^{i} = \mathcal{L}^{i}_{cla} + \varphi(t)\cdot\mathcal{L}^{i}_{ali},
\end{align}
where we employ a dynamic balancing function $w.r.t.$ the epoch number $t$.
We update the model by stochastic gradient descent without back-propagating towards the dashed arrow in Fig.~\ref{fig_pipeline}.
The pseudo-code is shown in Algorithm~\ref{alg:algorithm1}.

\section{Experiments}
\label{sec:experiments}

\subsection{Setup}
\paragraph{Datasets.}
We adopt four widely used benchmark datasets including Fashion-MNIST~\cite{xiao2017fashion}, SVHN~\cite{netzer2011reading}, CIFAR-10 and CIFAR-100~\cite{krizhevsky2009learning}.
We manually corrupt these datasets into partially labeled versions by uniform~\cite{wang2022pico} and instance-dependent~\cite{xu2021instance} generation process.
We further investigate PaPi on fine-grained image classification dataset CIFAR-100-H~\cite{wang2022pico} and few-shot image classification dataset Mini-Imagenet~\cite{vinyals2016matching}.
We set $q \in \{0.1, 0.3, 0.5, 0.7\}$ for Fashion-MNIST, SVHN, CIFAR-10 and CIFAR-100-H, set $q \in \{0.01, 0.05, 0.1, 0.2\}$ for CIFAR-100 and Mini-Imagenet.

\paragraph{Baselines.}
We compare PaPi against 7 state-of-the-art PLL methods:
(1) DPLL~\cite{wu2022DPLL} performs supervised learning on non-candidate labels and employs consistency regularization on candidate labels.
(2) PiCO~\cite{wang2022pico} identifies the true label by using contrastively learned representations.
(3) VALEN~\cite{xu2021instance} iteratively recovers the latent label distribution by adopting the variational inference.
(4) PRODEN~\cite{lv2020progressive} accomplishes the update of model and the identification of true labels in a seamless manner.
(5) LWS~\cite{wen2021leveraged} considers the trade-off between losses on candidate labels and non-candidate labels.
(6) RC~\cite{feng2020provably} utilizes the importance re-weighting strategy.
(7) CC~\cite{feng2020provably} employs a transition matrix to form an empirical risk estimator.

\paragraph{Implementation.}
For all datasets, we use an 18-layer ResNet following~\cite{he2016deep,wang2022pico} as the encoder $f(\cdot)$.
We instantiate $g(\cdot)$ as a 2-layer MLP with one hidden layer (as well as ReLU activation).
The classifier $h(\cdot)$ is a single linear layer, which is also used for inference.
We present results based on three independent runs for all methods.
Source code is available at https://github.com/AlphaXia/PaPi.

\begin{figure*}[ht]
  \centering
  \begin{subfigure}{0.245\linewidth}
    \includegraphics[scale=0.185]{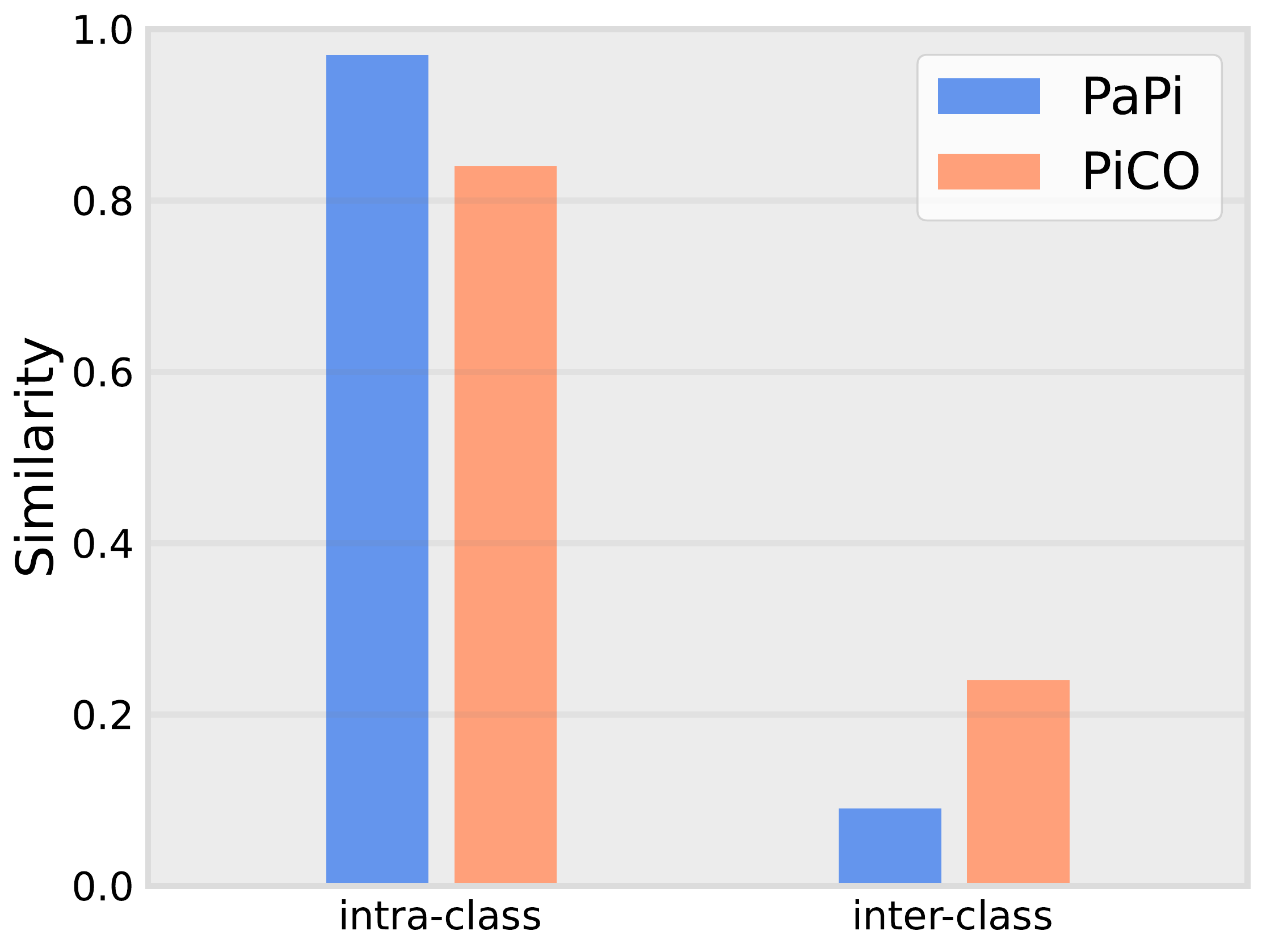}
    \caption{CIFAR-10 ($q$ = 0.5)}
    \label{fig:dis-1}
  \end{subfigure}
  \hfill
  \begin{subfigure}{0.245\linewidth}
    \includegraphics[scale=0.185]{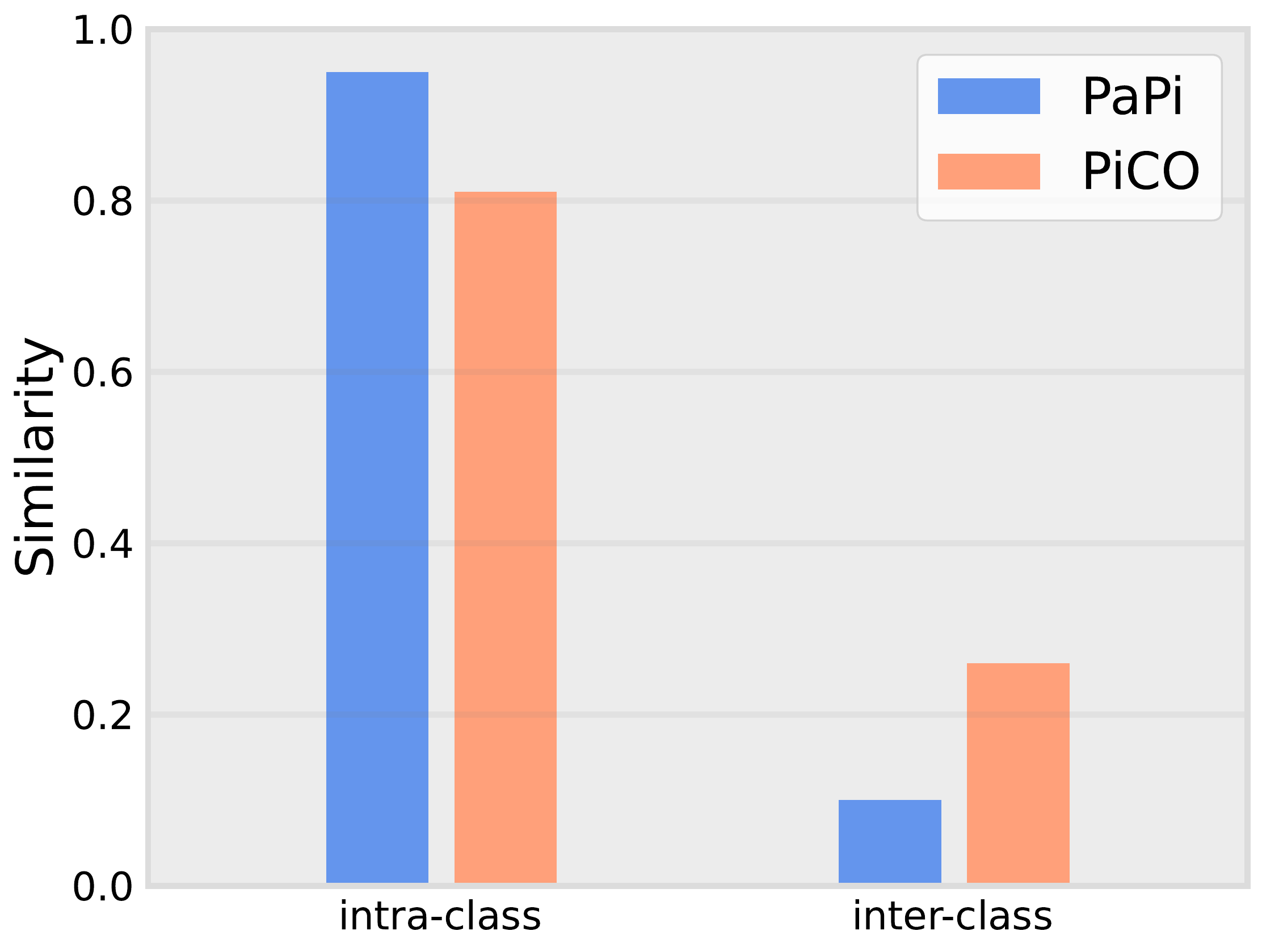}
    \caption{CIFAR-10 ($q$ = 0.7)}
    \label{fig:dis-2}
  \end{subfigure}
  \hfill
  \begin{subfigure}{0.245\linewidth}
    \includegraphics[scale=0.185]{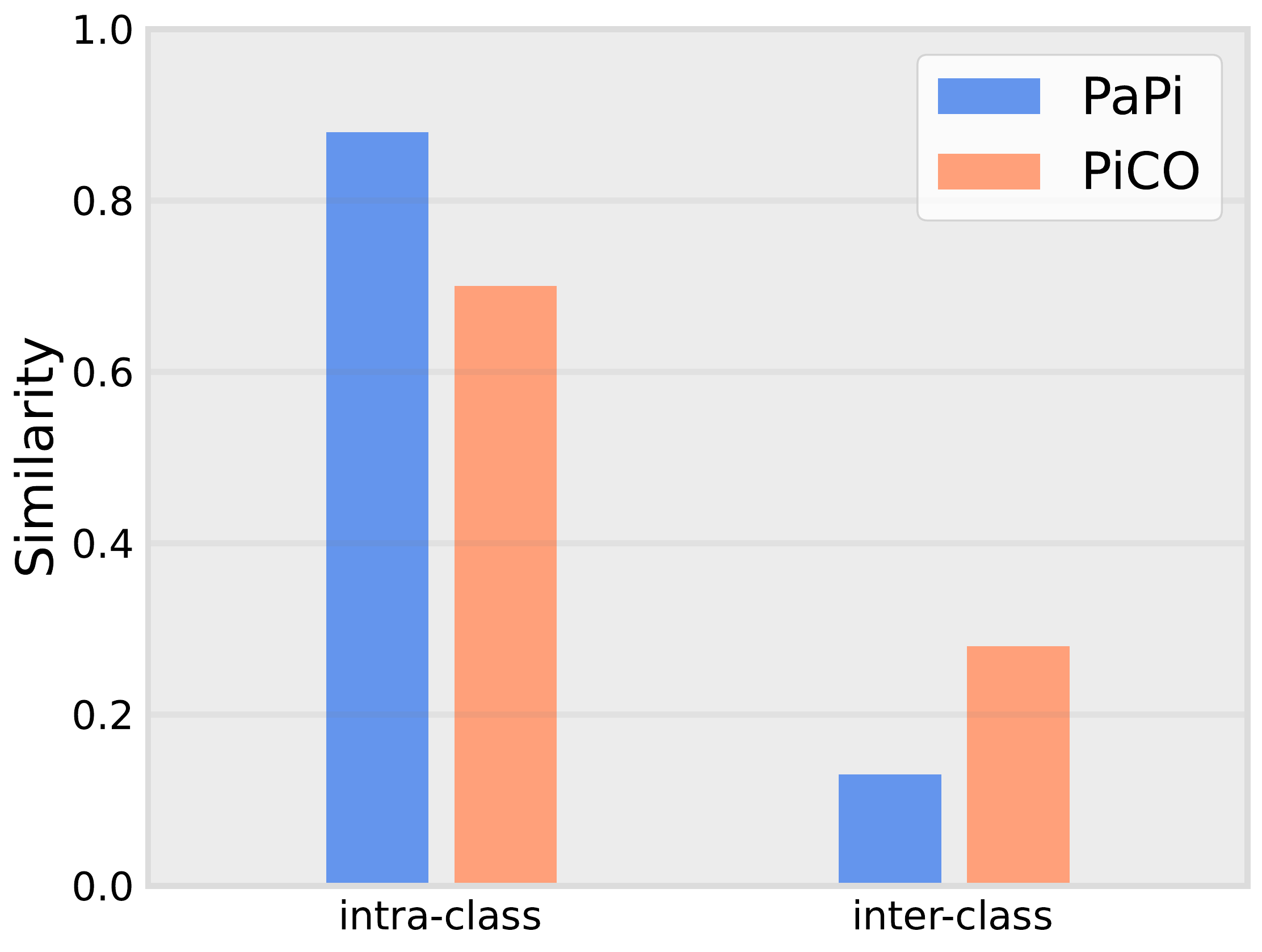}
    \caption{CIFAR-100 ($q$ = 0.1)}
    \label{fig:dis-3}
  \end{subfigure}
  \hfill
  \begin{subfigure}{0.245\linewidth}
    \includegraphics[scale=0.185]{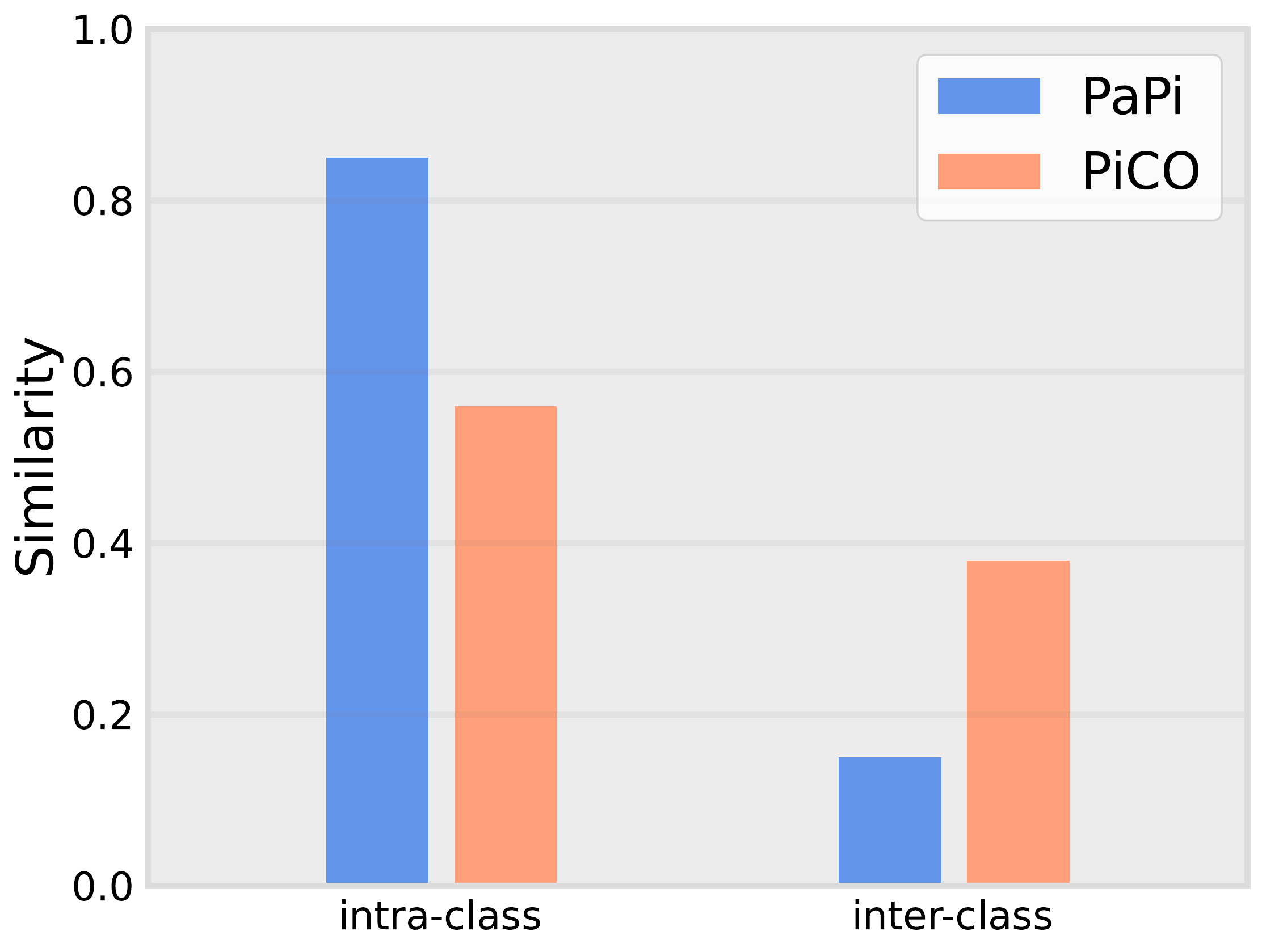}
    \caption{CIFAR-100 ($q$ = 0.2)}
    \label{fig:dis-4}
  \end{subfigure}

  \centering
  \begin{subfigure}{0.245\linewidth}
    \includegraphics[scale=0.185]{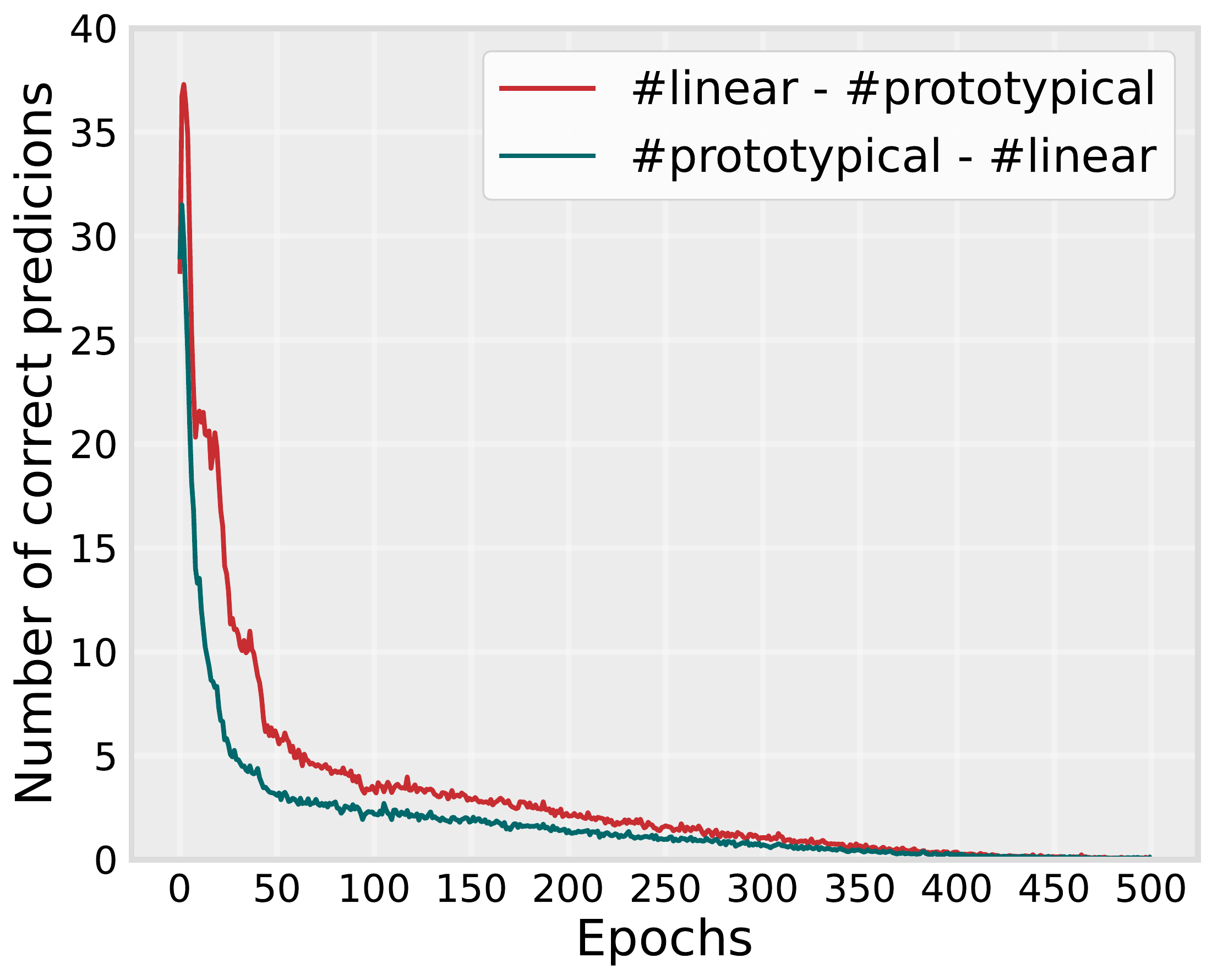}
    \caption{CIFAR-10 ($q$ = 0.5)}
    \label{fig:dir-1}
  \end{subfigure}
  \hfill
  \begin{subfigure}{0.245\linewidth}
    \includegraphics[scale=0.185]{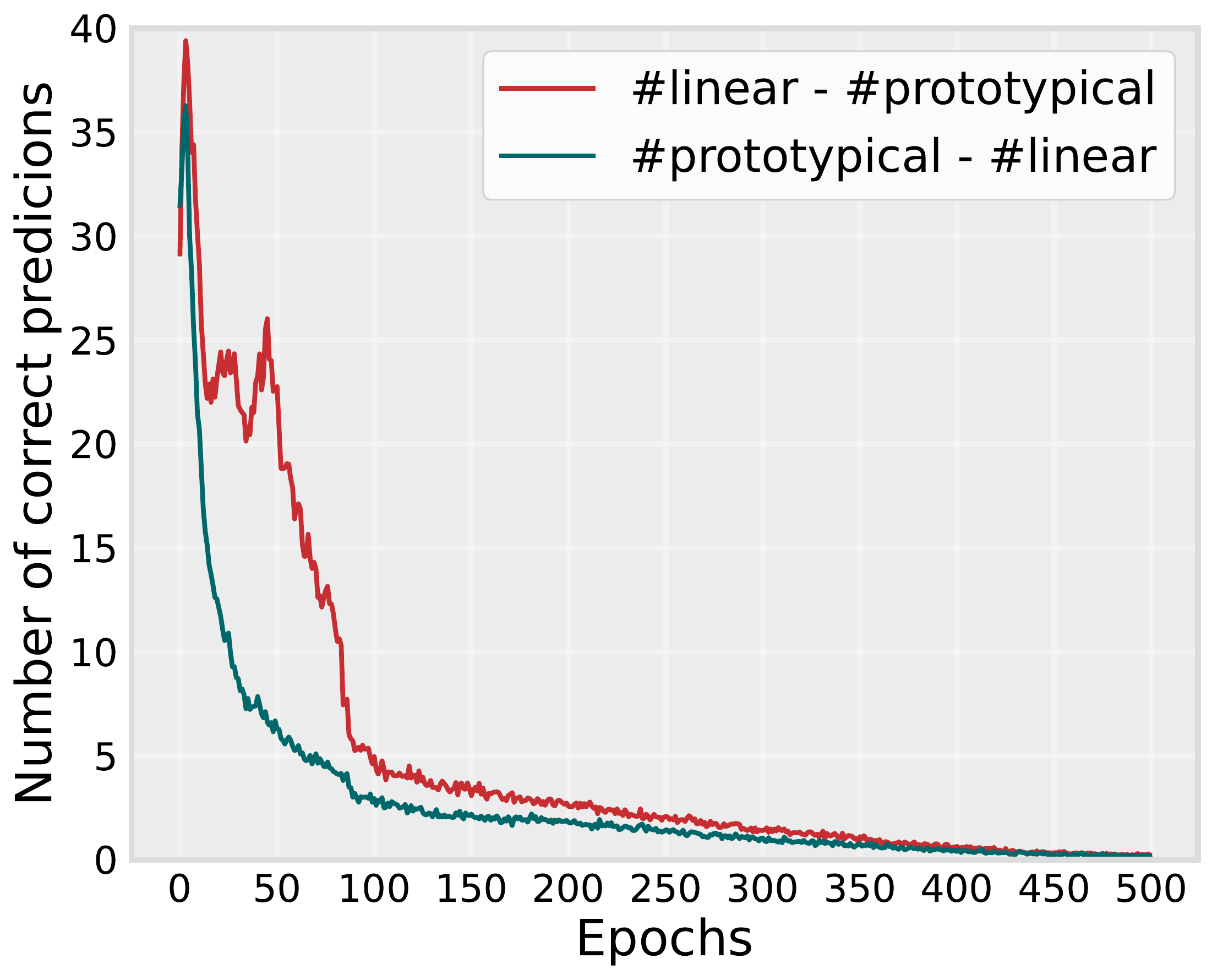}
    \caption{CIFAR-10 ($q$ = 0.7)}
    \label{fig:dir-2}
  \end{subfigure}
  \hfill
  \begin{subfigure}{0.245\linewidth}
    \includegraphics[scale=0.185]{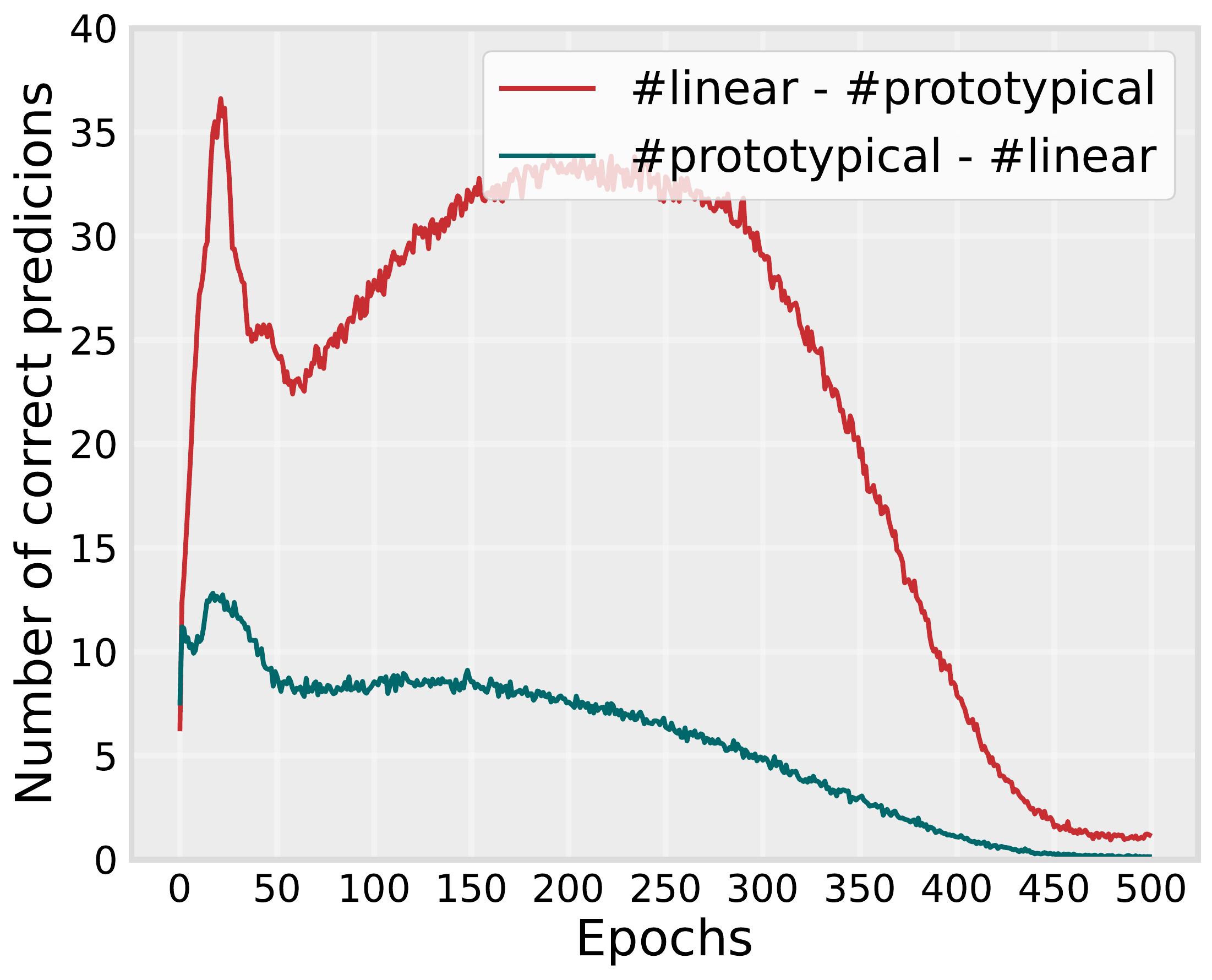}
    \caption{CIFAR-100 ($q$ = 0.1)}
    \label{fig:dir-3}
  \end{subfigure}
  \hfill
  \begin{subfigure}{0.245\linewidth}
    \includegraphics[scale=0.185]{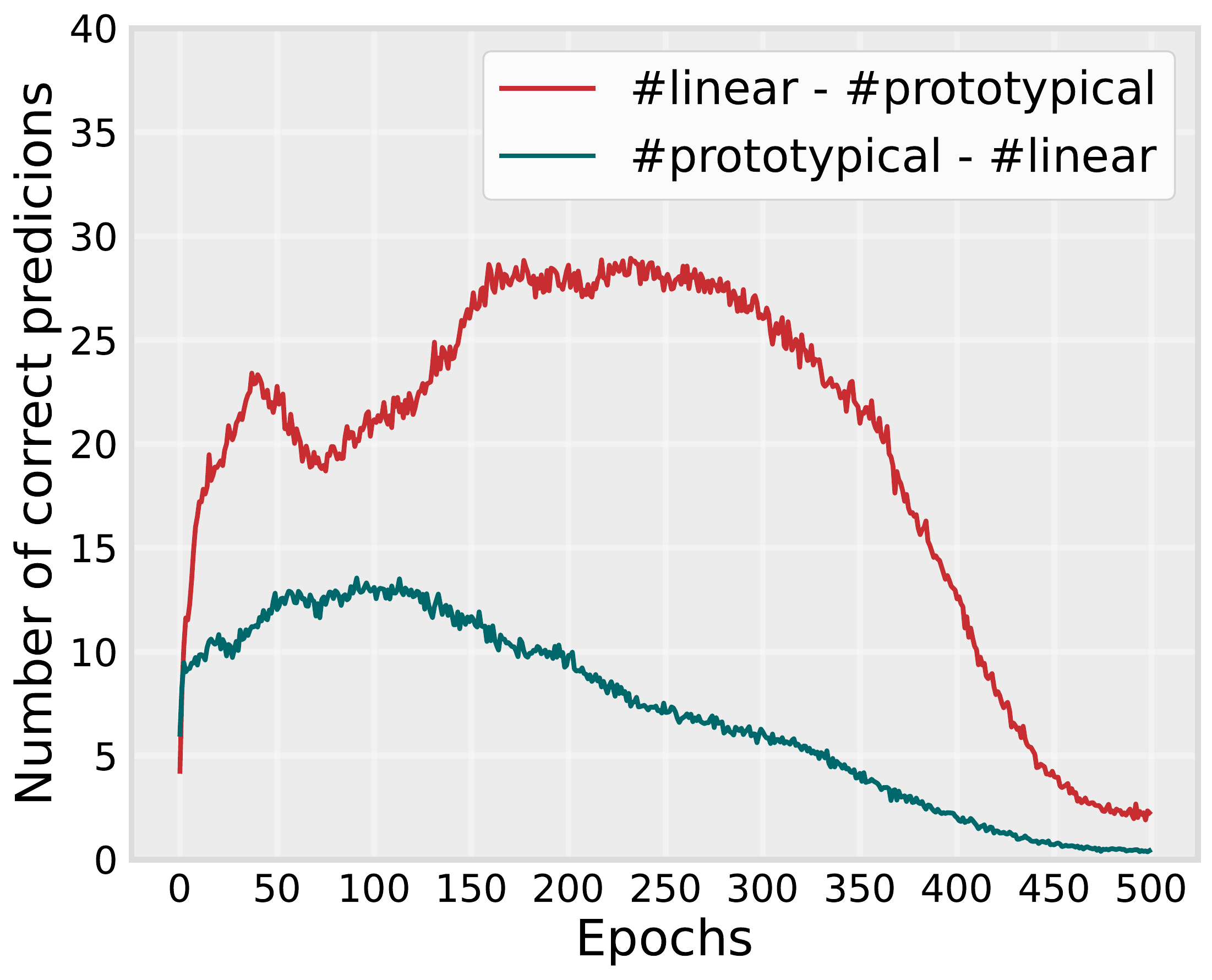}
    \caption{CIFAR-100 ($q$ = 0.2)}
    \label{fig:dir-4}
  \end{subfigure}

  \caption{Visualizations of the effectiveness of learned representation.
  (a)-(d): Intra-class similarity and inter-class similarity on CIFAR-10 ($q$ = 0.5, 0.7) and CIFAR-100 ($q$ = 0.1, 0.2). (e)-(h): The red lines indicate the number of samples that were correctly classified by the linear classifier and incorrectly classified by the prototypical classifier per mini-batch, and the green lines are the opposite.}
  \label{fig:effective_result1}
\end{figure*}

\begin{table*}[ht]
	\centering
	\setlength{\tabcolsep}{6.9mm}{
		\begin{tabular}{c|cccc}
			\toprule[1.5pt]
			Method & Fashion-MNIST & SVHN & CIFAR-10 & CIFAR-100\\
			\midrule
      PaPi (ours) & \textbf{93.84 $\pm$ 0.07\%} & \textbf{97.74 $\pm$ 0.05\%} & \textbf{95.13 $\pm$ 0.07\%} & \textbf{63.70 $\pm$ 0.19\%} \\
            DPLL & 92.72 $\pm$ 0.14\% & 92.11 $\pm$ 0.88\% & 90.73 $\pm$ 0.11\% & 33.14 $\pm$ 1.74\% \\
            PiCO & 87.86 $\pm$ 0.60\% & 91.18 $\pm$ 1.10\% & 90.50 $\pm$ 0.17\% & 59.13 $\pm$ 0.31\% \\
            VALEN & 88.89 $\pm$ 0.13\% & 94.66 $\pm$ 0.17\% & 90.16 $\pm$ 0.52\% & 29.25 $\pm$ 0.14\% \\
            PRODEN & 91.52 $\pm$ 0.24\% & 95.78 $\pm$ 0.46\% & 89.10 $\pm$ 0.44\% & 31.25 $\pm$ 0.20\% \\
            LWS & 71.68 $\pm$ 3.89\% & 54.49 $\pm$ 7.21\% & 48.67 $\pm$ 7.68\% & 25.14 $\pm$ 2.18\% \\
            RC & 91.42 $\pm$ 0.30\% & 94.85 $\pm$ 0.26\% & 88.66 $\pm$ 0.37\% & 38.25 $\pm$ 0.79\% \\
            CC & 91.43 $\pm$ 0.76\% & 95.08 $\pm$ 0.24\% & 88.33 $\pm$ 0.75\% & 43.30 $\pm$ 1.90\% \\
			\bottomrule[1.5pt]
		\end{tabular}
	}
	\caption{Classification accuracy (mean $\pm$ std) on Fashion-MNIST, SVHN, CIFAR-10 and CIFAR-100 with instance-dependent partial labels.
	The best accuracy is highlighted in bold.}
	\label{tab:instance_dependent_result1}
\end{table*}

\begin{figure*}[ht]
  \centering
  \begin{subfigure}{0.245\linewidth}
    \includegraphics[scale=0.185]{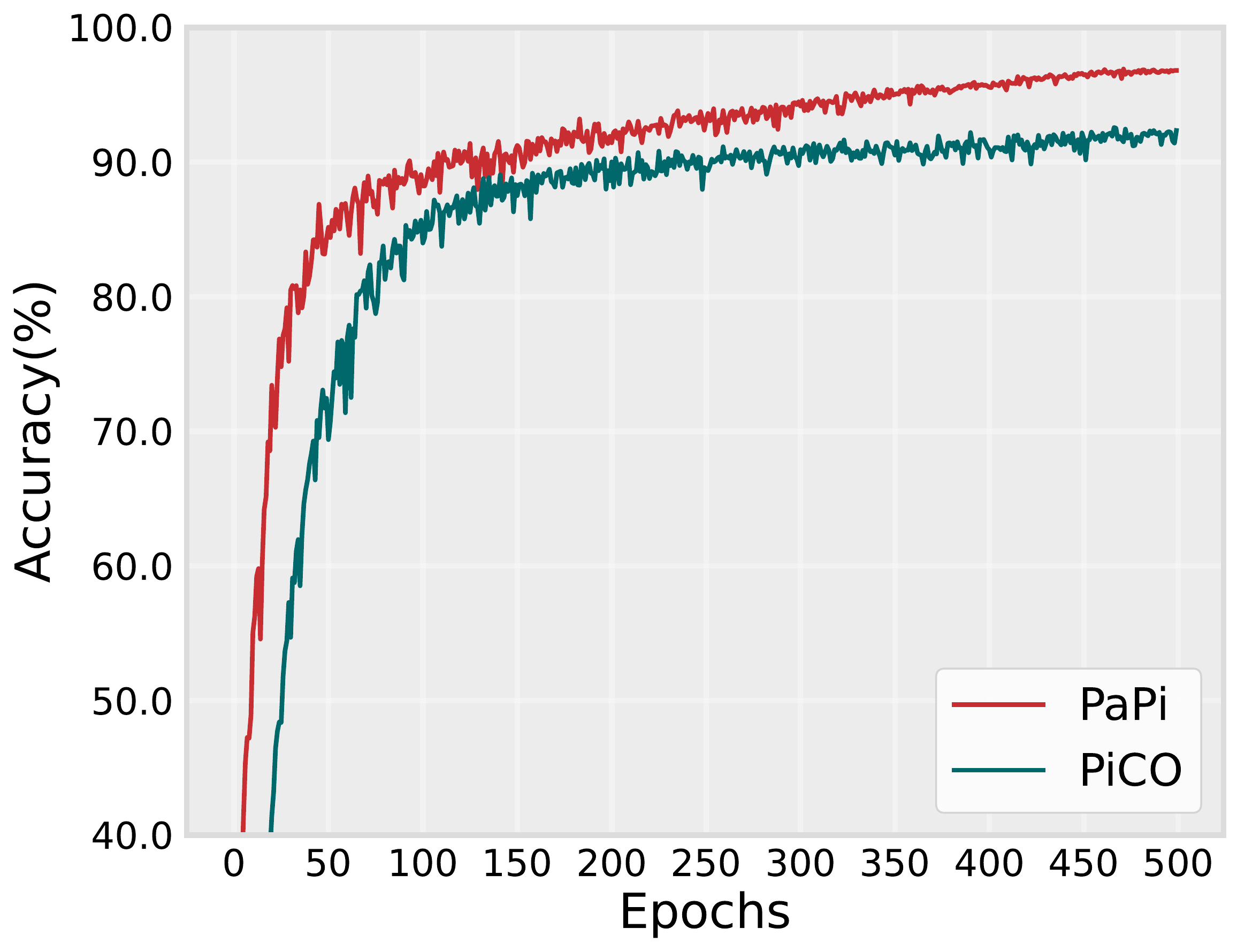}
    \caption{CIFAR-10 ($q$ = 0.5)}
    \label{fig:proto-1}
  \end{subfigure}
  \hfill
  \begin{subfigure}{0.245\linewidth}
    \includegraphics[scale=0.185]{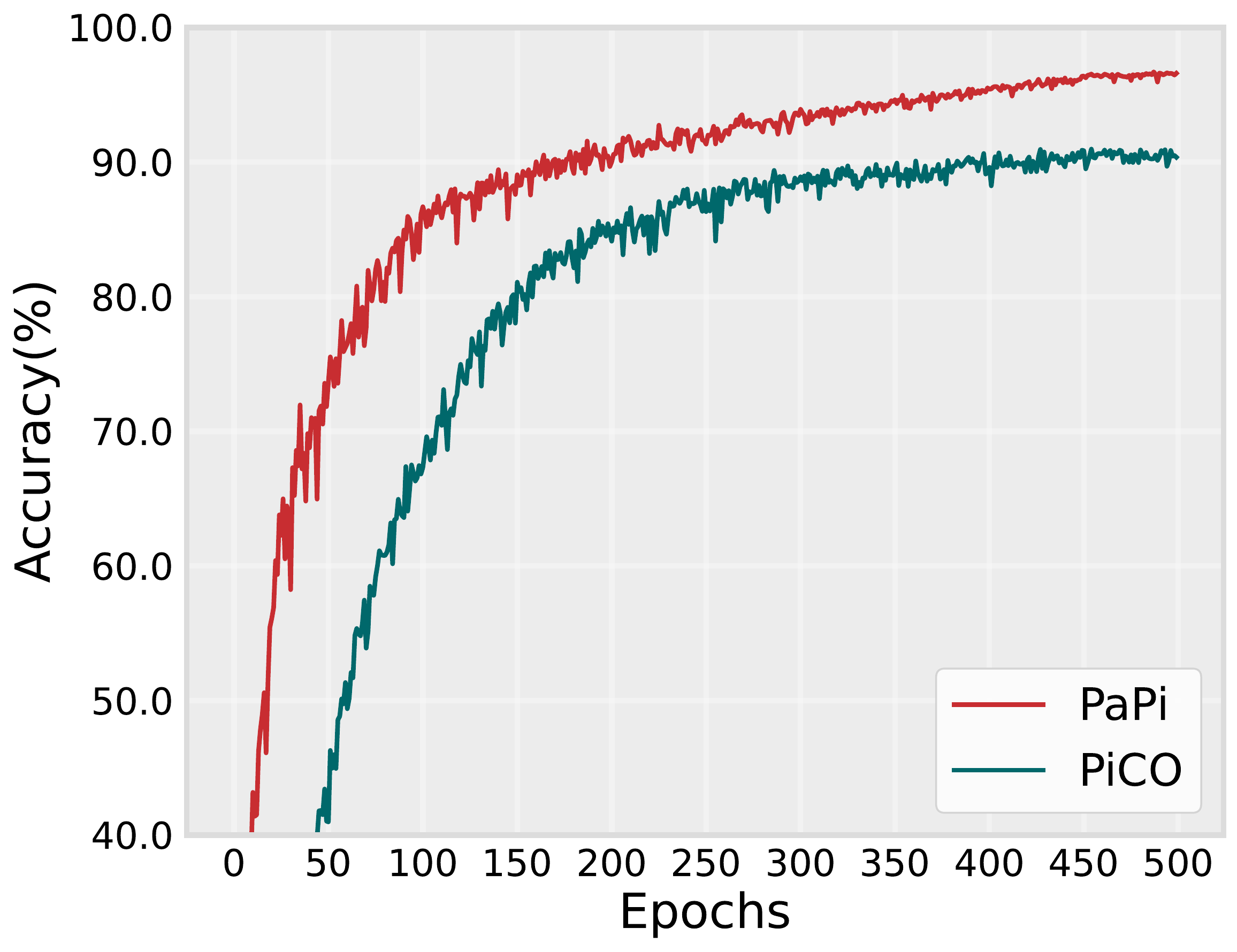}
    \caption{CIFAR-10 ($q$ = 0.7)}
    \label{fig:proto-2}
  \end{subfigure}
  \hfill
  \begin{subfigure}{0.245\linewidth}
    \includegraphics[scale=0.185]{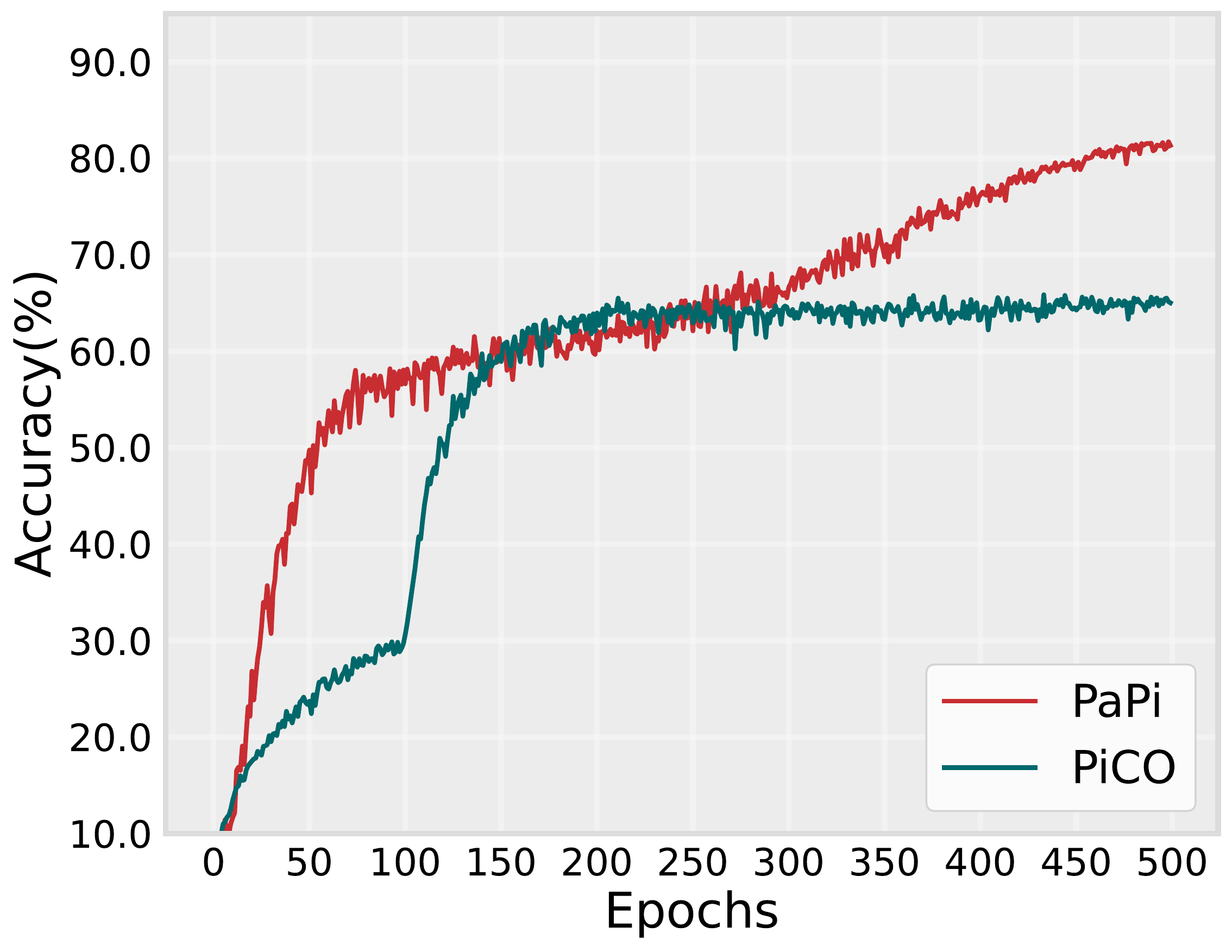}
    \caption{CIFAR-100 ($q$ = 0.1)}
    \label{fig:proto-3}
  \end{subfigure}
  \hfill
  \begin{subfigure}{0.245\linewidth}
    \includegraphics[scale=0.185]{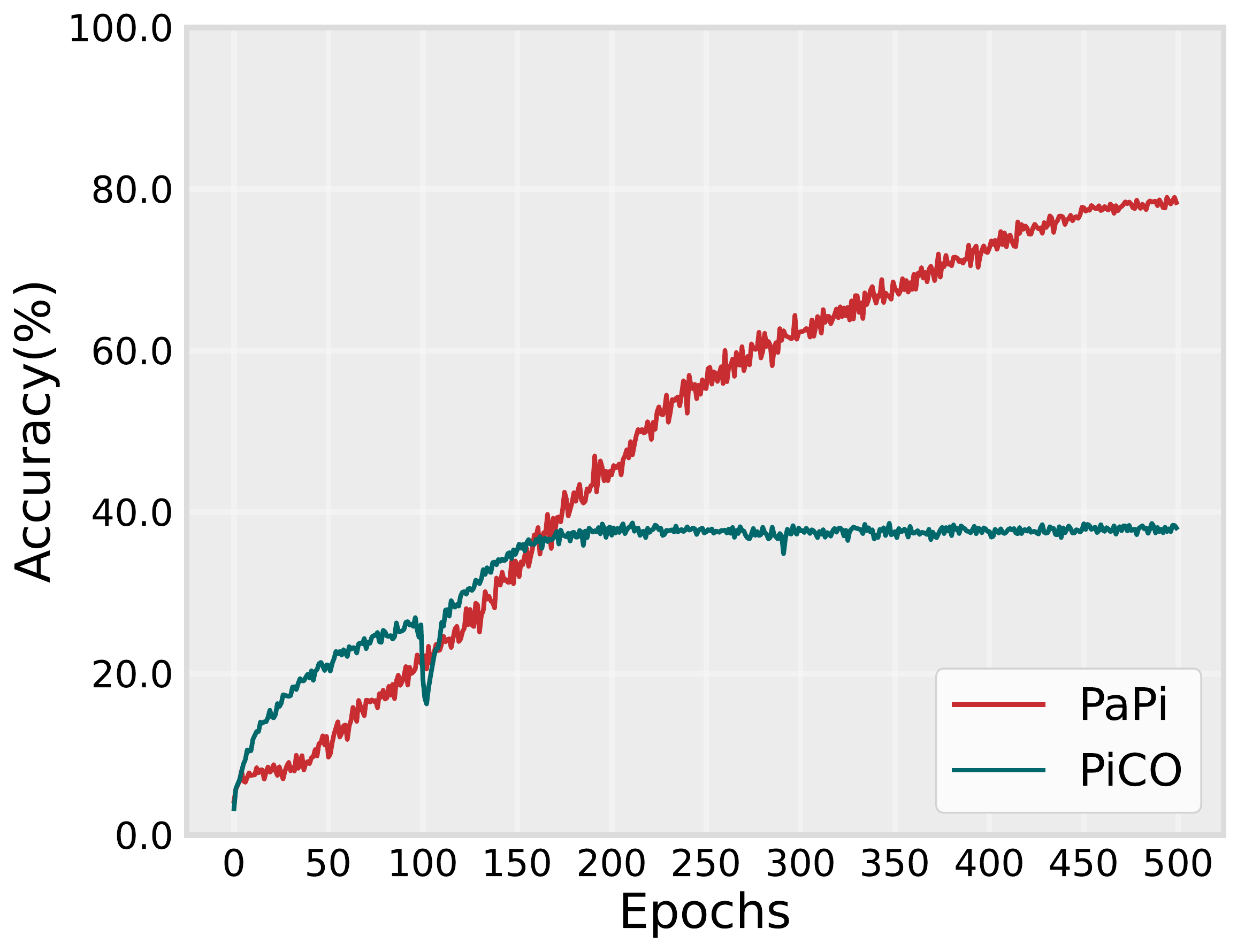}
    \caption{CIFAR-100 ($q$ = 0.2)}
    \label{fig:proto-4}
  \end{subfigure}

  \caption{Classification accuracy calculated with the learned prototypes on CIFAR-10 ($q$ = 0.5, 0.7) and CIFAR-100 ($q$ = 0.1, 0.2).}
  \label{fig:proto_result1}
\end{figure*}

\begin{table*}[ht]
	\centering
	\setlength{\tabcolsep}{7.0mm}{
		\begin{tabular}{c|cccc}
			\toprule[1.5pt]
			\multirow{2}*{Method} & \multicolumn{2}{c}{CIFAR-10} & \multicolumn{2}{c}{CIFAR-100} \\
			~ & $q$ = 0.5 & $q$ = 0.7 & $q$ = 0.1 & $q$ = 0.2 \\
			\midrule
		    PaPi (ours) & \textbf{96.90 $\pm$ 0.09\%} & \textbf{96.58 $\pm$ 0.07\%} & \textbf{81.65 $\pm$ 0.27\%} & \textbf{79.49 $\pm$ 0.22\%} \\
            \emph{Variant} 1 & 96.42 $\pm$ 0.11\% & 95.82 $\pm$ 0.12\% & 79.82 $\pm$ 0.09\% & 78.02 $\pm$ 0.05\% \\
            \emph{Variant} 2 & 94.39 $\pm$ 0.10\% & 92.26 $\pm$ 0.12\% & 77.48 $\pm$ 0.07\% & 72.30 $\pm$ 0.26\% \\
            PiCO & 93.97 $\pm$ 0.55\% & 92.88 $\pm$ 0.57\% & 71.55 $\pm$ 0.31\% & 48.76 $\pm$ 1.19\% \\
			\bottomrule[1.5pt]
		\end{tabular}
	}
	\caption{Classification accuracy (mean $\pm$ std) of different degenerated methods on CIFAR-10 ($q$ = 0.5, 0.7) and CIFAR-100 ($q$ = 0.1, 0.2).}
	\label{tab:variants_result1}
\end{table*}

\begin{table}[ht]
	\centering
	\setlength{\tabcolsep}{2.0mm}{
		\begin{tabular}{c|cccc}
			\toprule[1.5pt]
			\multirow{1}*{Data augmentation} & CIFAR-10 & CIFAR-100 \\
			\multirow{1}*{compositions} & $q$ = 0.5 & $q$ = 0.1 \\
			\midrule
		    PaPi (S+W) & 96.90 $\pm$ 0.09\% & \textbf{81.65 $\pm$ 0.27\%}  \\
            PaPi (2$\times$S) & \textbf{96.91 $\pm$ 0.05\%} & 80.96 $\pm$ 0.08\%  \\
            PaPi (2$\times$W) & 96.60 $\pm$ 0.03\% & 79.35 $\pm$ 0.14\%  \\
            \midrule
		    DPLL (2S+W) & 95.31 $\pm$ 0.11\% & 78.51 $\pm$ 0.24\%  \\
            DPLL (3$\times$S) & 95.67 $\pm$ 0.06\% & 78.41 $\pm$ 0.50\%  \\
            DPLL (3$\times$W) & 93.76 $\pm$ 0.21\% & 75.23 $\pm$ 0.24\%  \\
			\bottomrule[1.5pt]
		\end{tabular}
	}
	\caption{Accuracy (mean $\pm$ std) under different data augmentation compositions on CIFAR-10 ($q$ = 0.5) and CIFAR-100 ($q$ = 0.1).}
	\label{tab:da_result1}
\end{table}

\subsection{Main Results}
\paragraph{PaPi achieves sota results.}
We present results on Fashion-MNIST, SVHN and CIFAR-10 in Table~\ref{tab:benchmark_result1}.
Obviously, we can observe that PaPi significantly outperforms all baselines on all benchmark datasets, especially under \emph{high ambiguity levels}.
Specifically, PaPi outperforms current state-of-the-art method by $\bm{0.96\%}$ and $\bm{2.50\%}$ respectively on SVHN and CIFAR-10 when $q$ is set to 0.7.
More importantly, PaPi achieves comparable results against supervised learning model, showing that label disambiguation is satisfactorily completed in PaPi.
For example, PaPi suffers tiny performance degradation (lower than $\bm{1\%}$) compared with supervised learning model in many cases.
The performance degradation of PaPi is lower than $\bm{3\%}$ compared with supervised learning model \emph{even} under $q$ = 0.7, while most baselines demonstrate a significant performance drop.
In Table~\ref{tab:benchmark_result2}, we also observe that PaPi consistently outperforms all baselines on CIFAR-100 and the superiority is more substantial than that on the dataset with a small label space.
For example, PaPi outperforms the current state-of-the-art method by $\bm{3.14\%}$ and $\bm{3.72\%}$ respectively when $q$ is set to 0.1 and 0.2.

In a more practical and challenge PLL setting where the partial labels are generated in an instance-dependent fashion, PaPi also outperforms the current state-of-the-art method by a large margin.
As presented in Table~\ref{tab:instance_dependent_result1}, PaPi outperforms current state-of-the-art method by $\bm{4.40\%}$ on CIFAR-10 and $\bm{4.57\%}$ on CIFAR-100, which fully demonstrates the effectiveness of PaPi when facing more practical instance-dependent ambiguity.

From the above experimental results, we show that PaPi can effectively handle both uniform and instance-dependent PLL setting.
Besides, we investigate the performance of PaPi and all baselines on CIFAR-100-H and Mini-Imagenet, which are less explored in previous literature.
From Table~\ref{tab:benchmark_result1} and Table~\ref{tab:benchmark_result2}, we observe that PaPi still outperforms all baselines significantly.
The whole comparison results emphasize the superiority of our framework.

\paragraph{PaPi learns effective representations.}
We validate the effectiveness of learned representation by calculating the intra-class and inter-class similarity.
From Fig.~\ref{fig:dis-1} to Fig.~\ref{fig:dis-4}, we can observe that PaPi produces well-separated clusters and more distinguishable representations.
Moreover, we justify our correct disambiguation guidance direction.
From Fig.~\ref{fig:dir-1} to Fig.~\ref{fig:dir-4}, we find that the linear classifier always shows higher accuracy than the prototypical one and the linear classifier always has something new to teach the prototypical one until convergence.

\subsection{Ablation and Analysis}

\paragraph{The effectiveness of prototypical alignment term.}
We explore the effectiveness of our proposed prototypical alignment.
Specifically, we compare PaPi with several variants: (1) \emph{Variant} 1 which removes the mixup augmentation compared with PaPi; (2) \emph{Variant} 2 which removes the prototypical alignment loss compared with PaPi.
From Table~\ref{tab:variants_result1}, we observe that PaPi outperforms \emph{Variant} 1 (\eg, +$\bm{1.47\%}$ on CIFAR-100 ($q$ = 0.2)), which verifies the effectiveness of mixup augmentation.
We also observe that PaPi outperforms \emph{Variant} 2 (\eg, +$\bm{7.19\%}$ on CIFAR-100 ($q$ = 0.2)), which confirms the importance of prototypical alignment for identifying the true label.
Besides, we find that even \emph{Variant} 2 achieves comparable results against the current state-of-the-art method, which validates the importance of self-teaching fashion.

\paragraph{The performance of learned prototypical classifier.}
In Fig.~\ref{fig:proto_result1}, we report the classification accuracy calculated with the learned prototypes.
We can observe that PaPi achieves better performance especially facing \emph{high ambiguity levels}, which demonstrates the effectiveness of our learned prototypical classifier.
For example, PaPi outperforms PiCO by $\bm{6.24\%}$ and $\bm{40.29\%}$ respectively on CIFAR-10 ($q$ = 0.7) and CIFAR-100 ($q$ = 0.2).

\paragraph{The impact of data augmentation compositions.}
In our main experiments, we adopt one weak and one strong augmentation.
As shown in Table~\ref{tab:da_result1}, we compare the accuracy when equipped with different data augmentation compositions, where S means strong augmentation and W means weak augmentation.
We observe that the combination of strong and weak augmentation achieves the best performance.
Similarly, we evaluate DPLL~\cite{wu2022DPLL} under different compositions.
When equipped with two weak augmentations, PaPi suffers less performance drop compared with DPLL, which demonstrates the competitiveness of PaPi.

\section{Conclusion}
\label{sec:conclusion}
In this paper, we proposed a simple PLL framework termed PaPi which explicitly contrasts the prototypical similarity with the visual similarity between categories, such that PaPi is remarkable for improving the class-level discrimination of learned representation.
Extensive experimental results under multiple PLL settings demonstrated PaPi established new state-of-the-art performance especially when facing high ambiguity levels and instance-dependent ambiguity.

\paragraph{Acknowledgement.} 
This research was supported by the National Key Research \& Development Plan of China (No. 2018AAA0100104), the National Science Foundation of China (62125602, 62076063, 62206050), China Postdoctoral Science Foundation under Grant 2021M700023, Jiangsu Province Science Foundation for Youths under Grant BK20210220, the Young Elite Scientists Sponsorship Program of Jiangsu Association for Science and Technology TJ-2022-078, the Big Data Computing Center of Southeast University.

{\small
\bibliographystyle{ieee_fullname}
\bibliography{egbib}
}

\end{document}